\documentclass{article} 
\usepackage{preprint,times}
\usepackage{hyperref}
\usepackage{url}
\usepackage{multirow}
\PassOptionsToPackage{numbers,compress}{natbib}
\usepackage{xspace}
\usepackage[most]{tcolorbox}
\usepackage{algorithm}
\usepackage{algpseudocode}
\usepackage{pifont}

\usepackage{wrapfig}
\usepackage{makecell}
\usepackage{colortbl}
\usepackage{enumitem}
\usepackage{amsmath}
\usepackage{mdframed}
\usepackage{multirow}
\usepackage{fvextra}
\usepackage{silence}        
\usepackage[utf8]{inputenc} 
\usepackage[T1]{fontenc}    
\usepackage{hyperref}       
\usepackage{url}            
\usepackage{booktabs}       
\usepackage{amsfonts}       
\usepackage{nicefrac}       
\usepackage{microtype}      
\usepackage{graphicx}
\usepackage{multirow}
\usepackage{tikz}
\usepackage{subcaption}     

\newcommand{\lei}[1]{}
\newcommand{\dmh}[1]{}
\newcommand{\josh}[1]{}
\newcommand{\xu}[1]{}
\newcommand{\pmv}[1]{}
\newcommand{\lcao}[1]{}
\newcommand{\ear}[1]{}

\usepackage{amsmath,amsfonts,bm}

\def\eqref#1{equation~\ref{#1}}

\def\1{\bm{1}}

\DeclareMathAlphabet{\mathsfit}{\encodingdefault}{\sfdefault}{m}{sl}
\SetMathAlphabet{\mathsfit}{bold}{\encodingdefault}{\sfdefault}{bx}{n}

\newcolumntype{C}[1]{>{\centering\arraybackslash}p{#1}}
\definecolor{fullgreen}{HTML}{008A00}
\colorlet{halfgreen}{fullgreen!15}

\definecolor{bulbyellow}{RGB}{247,208,70}

\newcommand{\costA}{{\color{green!50!black}\$\$\$\$\$\$}}
\newcommand{\costB}{{\color{green!50!black}\$\$\$\$\$}}
\newcommand{\costC}{{\color{green!50!black}\$\$\$\$}}
\newcommand{\costD}{{\color{green!50!black}\$\$\$}}
\newcommand{\costE}{{\color{green!50!black}\$\$}}
\newcommand{\costF}{{\color{green!50!black}\$}}

\definecolor{hlblue}{HTML}{D2E2F5}
\definecolor{hlorange}{HTML}{FFE0C0}
\newcommand{\app}{\textsc{MaadBench}}
\newcommand{\appeval}{\textsc{MaadBench-Eval}}
\newcommand{\appfull}{\textsc{MaadBench-Full}}

\newcommand{\eg}{\textit{e.g.},\xspace}
\newcommand{\ie}{\textit{i.e.},\xspace}
\newcommand{\etc}{\textit{e.t.c}\xspace}

\newcommand{\appfullurl}{\url{https://huggingface.co/datasets/hww123/MAADBench-full}}

\newcommand{\taskspace}{$10^{37}$}

\newcommand{\fullsize}{5,200}

\DeclareFontShape{T1}{ptm}{m}{scit}{<-> ssub * lmr/m/scsl}{} 
\title{\app{}: The Refreshable Paradigm for Anomaly Detection in Multi-Agent Systems}

\author{
\textbf{
Lei Ma$^{1}$,
Dennis Hofmann$^{1}$,
Haowen Xu$^{1}$,
Joshua DeOliveira$^{1}$,
Peter VanNostrand$^{1}$
}
\\[2pt]
\textbf{
Lei Cao$^{2}$,
Elke Rundensteiner$^{1}$
}
\\[6pt]
$^{1}$Worcester Polytechnic Institute
\\
$^{2}$University of Arizona
\\[4pt]
\texttt{\{lma5,dmhofmann,hxu4,jcdeoliveira,pvannostrand,rundenst\}@wpi.edu}
\\
\texttt{caolei@arizona.edu}
}

\iclrfinalcopy
\begin{document}
\maketitle

\pagestyle{fancy}
\fancyhead{}
\renewcommand{\headrulewidth}{0pt}

\maketitle
\begin{abstract}
    Recent studies report LLM-based multi-agent systems (MAS) fail at rates of 41\%–87\%, yet to our knowledge, no benchmark to date supports systematic anomaly detection (AD) for them. Building MAS AD benchmarks is hard because they must remain valid as LLM systems evolve:
    tasks may leak into training data and thus be memorized by LLMs, traces and anomaly patterns expire as backbones evolve, and labels must be provided reliably for each refresh.
    To address these challenges, we present \textbf{\app{}} (\textbf{MA}: multi-agent; \textbf{AD}: anomaly detection), the first refreshable MAS AD benchmark designed for diverse evolving LLM backbones underlying the agents. 
    \app{} combines (1) sampled-and-coupled generative tasks over a $\approx$\taskspace{}-task space to mitigate task leakage, (2) refreshable trace generation under configurable LLM backbones, and (3) automated provision of cost-free, deterministic step-level labels for fine-grained AD evaluation.
    Beyond offering the paradigm itself, we run \app{} with five SOTA LLM backbones, and release the dataset \textbf{\appfull{}} with \fullsize{} step-labeled traces. Benchmarking 25 AD methods on  \app{}  dataset  reveals substantial limitations in current approaches: they rely heavily on supervision, struggle with subtle MAS-specific anomalies, and lack robustness across LLM backbones. These gaps point to a rich research agenda for MAS-specific anomaly detection, with \app{} providing a systematic and refreshable testbed for method development and evaluation. We open-source \appfull{} at \appfullurl{}. 
\end{abstract}


\section{Introduction}





LLM-based multi-agent systems (MAS) have emerged as a promising solution for complex
collaborative tasks~\citep{liang2024encouraging, 10903668,gemma4_deepmind_2026}, 
but their reliability remains a major concern.
Recent studies report failure rates of 41\%--87\% even in state-of-the-art frameworks~\citep{MAST}.
Prior works study MAS reliability through failure analysis~\citep{MAST,Trail,whoandWhen,AgenTracer} and robustness mechanisms~\citep{gsafeguard,BlindGuard,yang2024watch}, both naturally connected to anomaly detection (AD) by identifying deviations from expected MAS execution. While AD has been extensively studied in traditional settings~\citep{chandola2009anomaly,pang2021deep,ADBench}, it remains underexplored for MAS. This motivates us to evaluate established AD techniques for detecting MAS errors and identify gaps requiring MAS-tailored detectors. 

MAS anomaly detection remains underexplored, in part because MAS execution and data traces differ fundamentally from traditional AD targets. MAS traces are textual outputs from evolving, multi-step execution processes~\citep{wu2024autogen,langgraph2024}, rather than fixed generate-once instances such as images or tabular records. 
They 
correspond
to
open-vocabulary natural language outputs, where anomalies tend
to be semantic, logic-rich, arising from reasoning and inter-agent 
communication errors~\citep{MAST, gsafeguard}, rather than numerical or categorical deviations as common
in system telemetry.

\begin{figure}[t]
    \centering
    \includegraphics[width=0.95\linewidth]{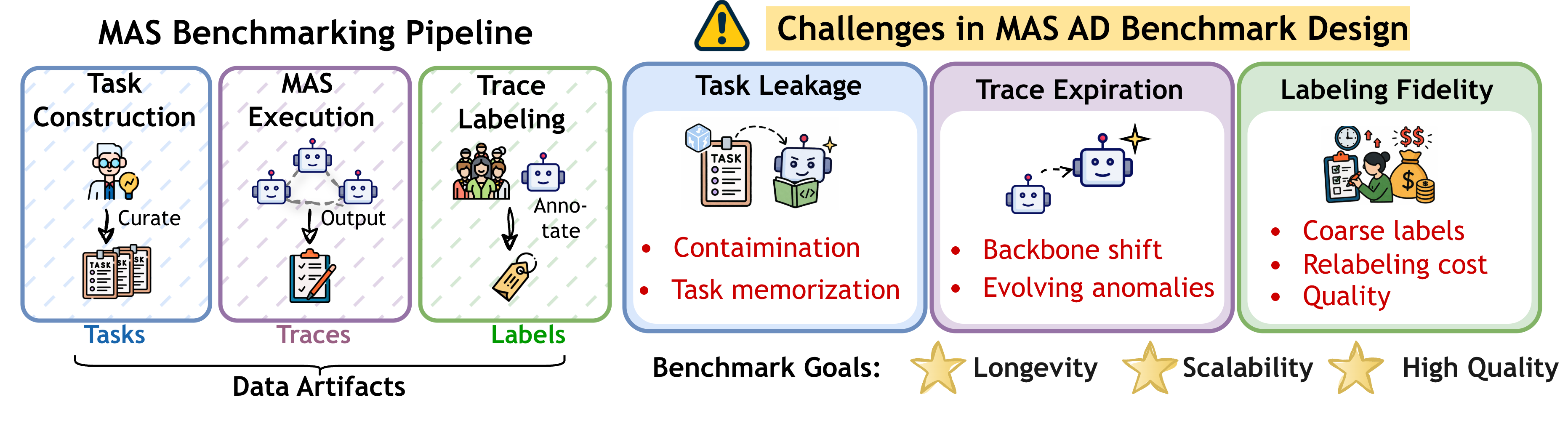}
    \vspace{-10pt}
\caption{
Overview of the MAS AD benchmarking pipeline (left) and the key design challenges (right) that must be addressed to achieve a long-lasting, scalable, and credible MAS AD benchmark.
}
    \label{fig:position}
    \vspace{-15pt}
\end{figure}

This introduces new design challenges unique to MAS AD 
benchmarks. As illustrated in Figure~\ref{fig:position}, a 
MAS benchmarking pipeline produces 
tasks and traces, and ideally also
the 
corresponding
labels  as benchmark artifacts. Static MAS benchmark artifacts, however, can rapidly expire as the 
underlying LLMs plugged in
for the agents
continue to evolve at an unprecedented speed. 
%
To ensure the benchmark's 
longevity, 
scalability and high quality, we must address the following three 
challenges:



\textbf{Challenge 1: Task Leakage.}
Benchmark leakage may cause LLMs to
memorize answers rather than reason genuinely~\citep{magar2022contamination,golchin2024timetravelllms,deng2024survey,deng2023investigating,xu2024benchmarking}. 
For a MAS AD benchmark, this is especially harmful, as it may fail to capture authentic anomaly patterns that constitute the benchmark’s core artifact.

\textbf{Challenge 2: Trace Expiration.}
MAS traces and anomalies reflect the specific LLM backbone and version utilized to  generate them~\citep{agentbench,chen2023chatgptsbehaviorchangingtime}. As LLMs evolve,
their tackling of tasks may
 yield different traces and anomaly patterns, causing the prior collected traces to expire.

 \textbf{Challenge 3: Label Fidelity.}
Useful MAS labels must be fine-grained, reproducible, and high-quality, but achieving all three is difficult. Fine-grained step-level labels provide stronger supervision~\citep{prm800k,math_shepherd} yet require costly inspection of long multi-agent traces. Worse, as traces are refreshed under evolving backbones, labels must also be regenerated on the fly. Balancing efficient label refresh with labeling cost and quality remains a key challenge. 


\begin{table*}[h]
\centering
\vspace{-5pt}
\caption{Comparison of SOTA MAS trace-analysis benchmarks across three core challenges. \app{} addresses all three to support systematic MAS AD development and evaluation.}
\label{tab:benchmark_comparison}
\resizebox{1.0\linewidth}{!}{
\begin{tabular}{lccccccc}
\toprule
\multirow{2}{*}{\textbf{Benchmark}} 
& \multicolumn{2}{c}{\textbf{CH1: Task Leakage}}
& \textbf{CH2: Trace}
& \multicolumn{3}{c}{\textbf{CH3: Labeling Fidelity}} 
& \textbf{AD} \\
\cmidrule(lr){2-3}
\cmidrule(lr){5-7}
& Generative 
& Space~$^\dagger$
& \textbf{Expiration} 
& Step-level 
& Relabeling Cost~$\S$ 
& Quality Guarantee 
& \textbf{Support}\\
\midrule \midrule
MAST~\citep{MAST}        
& \texttimes{} (fixed) 
& 1,642 
& \texttimes{} (non-refresh.) 
& \texttimes{} (trace) 
& \costC
& \texttimes{} (human/LLM) 
& Limited \\

TRAIL~\citep{Trail}       
& \texttimes{} (fixed) 
& 148 
& \texttimes{} (non-refresh.) 
& \texttimes{} (trace) 
& \costA
& \texttimes{} (human/LLM) 
& Limited \\

Silent Failures*~\citep{SilentFailures}
& \texttimes{} (fixed) 
& 5,169 
& \texttimes{} (non-refresh.) 
& \texttimes{} (trace) 
& \costD
& \texttimes{} (human) 
& Limited \\

Who\&When~\citep{whoandWhen}  
& \texttimes{} (fixed) 
& 184 
& \texttimes{} (non-refresh.) 
& \checkmark{} (step) 
& \costB  
& \texttimes{} (human/LLM) 
& Limited \\

AgenTracer~\citep{AgenTracer}
& \texttimes{} (fixed) 
& 2,500+ 
& \texttimes{} (non-refresh.) 
& \checkmark{} (step) 
& \costE
& \texttimes{} (LLM-assisted) 
& Limited \\

\midrule
\textbf{\app{} (ours)} 
& \textbf{\checkmark{} (generative)} 
& \textbf{$\approx$\taskspace{}} 
& \textbf{\checkmark{} (refresh.)} 
& \checkmark{} (\textbf{step}) 
& \textbf{\costF} 
& \textbf{\checkmark{} (deterministic)} 
& \textbf{Full} \\
\bottomrule
\end{tabular}}

\raggedright\small
\textsuperscript{*}Dataset not yet publicly released.
\textsuperscript{\S} Relabeling cost per trace (Appendix~\ref{app:label_cost}).
\textsuperscript{$\dagger$} Task space: prior benchmarks limited to released static traces, while our \app{} expands space through generation (Appendix~\ref{app:instance-space}).
\vspace{-5pt}
\end{table*}


\noindent\textbf{State-of-the-Art Benchmarks on MAS Trace Analysis.}
We characterize existing MAS trace-analysis benchmarks in Table~\ref{tab:benchmark_comparison}. 
Recent works have built MAS trace benchmarks by collecting and annotating traces from deployed systems~\citep{MAST, Trail, whoandWhen, AgenTracer,SilentFailures}.
These efforts include \textit{MAST}~\citep{MAST} for failure taxonomies, \textit{TRAIL}~\citep{Trail} for error localization, \textit{Who\&When}~\citep{whoandWhen} for agent/step attribution, and \textit{AgenTracer}~\citep{AgenTracer} for automated annotation via replay or fault injection.
\textit{Silent Failures}~\citep{SilentFailures} is closest in spirit to MAS AD, providing two labeled datasets for anomaly detection, but neither has been publicly released. 
While valuable, these benchmarks remain tied to fixed task sets (\textit{CH 1}), static trace collections (\textit{CH 2}), and one-time human or LLM labeling tied to specific backbones (\textit{CH 3}), limiting their use as evolving infrastructures for MAS AD development and evaluation. 
A comprehensive survey of related works 
is in Appendix~\ref{app:related_work}.


\textbf{\app{}: The Refreshable Benchmarking Paradigm for MAS AD.}
\app{} 
addresses the three core challenges of MAS AD benchmarking. To address \emph{task leakage}, it samples and couples hidden-dependency atomic tasks at configurable complexity levels, yielding a huge combinatorial task space rather than a fixed set. To handle \emph{trace expiration}, it enables trace refreshability by executing 
the tasks 
under new LLM backbones.
To ensure \emph{label fidelity}, each subtask is generated with oracle ground truth, enabling 
automatic deterministic step-level labeling without requiring human or LLM annotation.
Further,
\app{} elicits \textit{diverse MAS capabilities} 
across domains, including math, coding, planning, reasoning, and communication, with multiple task difficulties,  producing rich anomaly types for comprehensive AD evaluation. 



Our contributions are summarized as follows:

\begin{enumerate}[leftmargin=*, itemsep=0pt, topsep=0pt]

\item \textbf{\app{}: The Refreshable Benchmarking Paradigm for MAS AD.}
We establish principles for refreshable MAS AD benchmarking, realize them through bottom-up task construction, and instantiate them in \app{} to address task leakage, trace expiration, and label fidelity.

\item 
\textbf{Multi-LLM-as-Backbone MAS AD 
Datasets.} We run \app{} with five LLM backbones and multiple configurations to generate and release \appfull{}, containing \fullsize{} MAS traces 
with 
deterministic step-level labels.


\item \textbf{Broad Evaluation Across AD Methods and LLM Backbones.}
We profile MAS anomalies across action types, LLM backbones, observability, and propagation. We evaluate 25 AD methods across diverse supervision levels and detector families, spanning tabular, graph-based, MAS-specific, and LLM-as-a-judge approaches. 
This evaluation 
studies their
performance, robustness across anomaly profiles, and propagation capture.


\item \textbf{New Findings on MAS Anomaly Detection.} Our evaluation shows that the MAS AD problem
is
hard:
SOTA AD
methods on MAS 
rely heavily on supervision, with a $\sim1.5\times$ larger gap than tabular 
AD~\citep{ADBench}. Their performance varies across LLM backbones, while the long-horizon reasoning, task-failing, and propagation-origin anomalies remain hardest to detect. These findings outline a research agenda for MAS-tailored 
AD that prioritizes less supervision, 
backbone robustness, semantic understanding, and root-cause awareness.
\end{enumerate}

\section{\app{} Construction and Instantiation}
\label{sec:task}
We now present the design principles and realization of \app{}. A MAS AD benchmark should satisfy two goals: (1) complex tasks that elicit multi-agent coordination and diverse anomalies, and (2) automated, fine-grained ground-truth labeling of anomalous agent behaviors. Existing benchmarks~\citep{MAST,Trail,AgenTracer,whoandWhen} use complex collaborative tasks but leave task decomposition and execution to the MAS and annotate traces afterward. This separation of task construction, execution, and labeling makes automated generation of reliable step-level labels difficult.


\textbf{Bottom-up Oracle-Guided Construction.}
To support diverse and complex tasks while retaining deterministic fine-grained labels, we construct the benchmark bottom-up from atomic tasks, all of which are equipped 
with verified ground truth. This bottom-up design allows atomic tasks from different domains and benchmarks to be flexibly plugged into the framework. 
Atomic tasks are composed through \textit{oracle dependencies}, where the output of one task may serve as a prerequisite or input to another. Given task ground truth and oracle dependencies, labels can be deterministically propagated along intra-level input-output dependencies and across higher levels of the task hierarchy. 
\textit{
This principle enables large-scale combinatorial task generation with flexible configuration yet with deterministic labels at 
multiple task levels.}

\begin{figure}[t]
    \centering
    \includegraphics[width=0.99\linewidth]{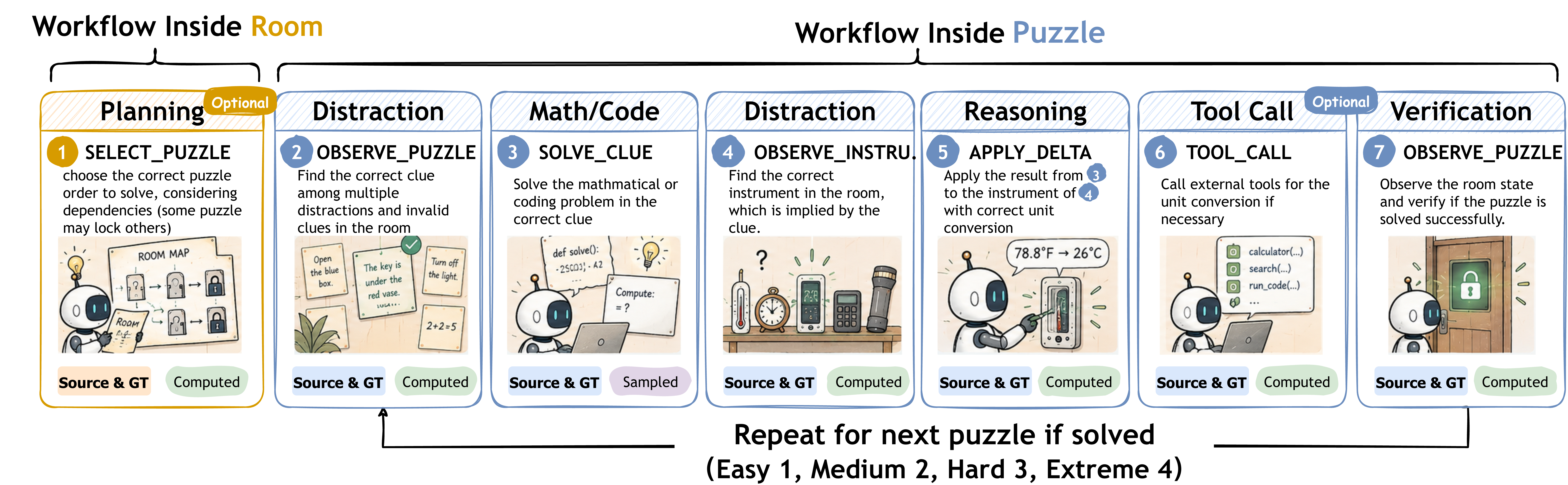}
        \vspace{-5pt}
    \caption{\textbf{MAADBench  Instantiation and MAS Execution.}
    \textbf{1) Workflow: } The execution workflow follows the same hierarchy: at the room level, the MAS resolves cross-puzzle dependencies; at the puzzle level, it executes atomic tasks with input-output dependencies. 
    \textbf{2) Top-level Composite Task (Room):} The MAS first needs to reason about the correct puzzle execution order under cross-puzzle dependencies, as a planning task. 
    \textbf{3) Intermediate Composite Task (Puzzle): } Each selected puzzle needs to be solved through a sequence of interdependent agent actions examining different capabilities. The \textsc{Solve\_Clue} tasks are sampled from GSM-Hard (Math) and LiveCodeBench-Execution (Code), while the remaining actions are generated from parameterized templates with algorithmic ground truth. 
    \textbf{4) Overall Task Difficulty:} We define four room difficulty levels: Easy, Medium, Hard, and Extreme, by increasing the number of puzzles and the complexity of their cross-puzzle dependencies. 
    }
    \vspace{-10pt}
    \label{fig:puzzle}

\end{figure}

\textbf{An Instantiation of \app{}.}
The atomic task space of \app{} spans diverse capabilities, including mathematical and code reasoning, planning, distraction handling, environment interaction, tool use, and verification. They can be sampled from current benchmarks with deterministic ground truth (e.g., LiveCodeBench-Execution~\citep{jain2025livecodebench}, GSM-Hard~\citep{GSM-hard}, GSM-Symbolic~\citep{GSM-Symbolic}, \etc), extended to future benchmarks as LLM capabilities evolve, or generated from parameterized templates with algorithmic ground truth. To compose these atomic tasks into a meaningful MAS scenario, we use an escape room as a \textit{metaphor for task composition}: each agent \textit{action} executes an atomic task; atomic tasks (\textit{actions}) are composed into intermediate composite tasks (\textit{puzzles}), which are further composed into top-level composite tasks (\textit{rooms}). The design supports configurable task counts and dependency complexity across difficulty levels, yielding diverse task instances that mitigate memorization and task leakage. Figure~\ref{fig:puzzle} summarizes one concrete instantiation of \app{}, including its task configuration, capabilities, and ground-truth provenance; full details and a room example are provided in Appendix~\ref{app:room_setup}.

\textbf{MAS Workflow and Trace Collection.}
Following the task design above, we build a multi-agent system where specialized agents execute the workflow in Figure~\ref {fig:puzzle} and communicate through structured JSON messages containing intermediate reasoning and outputs. We record the acting agent, action type, input/output messages, and rich runtime telemetry (e.g., token usage, latency, \etc). Additional implementation details on the MAS harness and tool calls are provided in Appendix~\ref{app:mas}.

\color{black}



\section{Profiling MAS Anomalies: Actions, Backbones, Observability, and Propagation}
\label{sec:execution}

Building on the above design, we now deploy the \app{} framework to generate a sample dataset, i.e.,  \appfull{}, and use it to profile 
the emergent MAS anomalies in terms of LLM capabilities, LLM backbones, observability, 
and propagation patterns. \textit{Note \app{} provides automatic mapping from 
anomalies to the MAST~\citep{MAST} failure taxonomy}, 
enabling benchmark labels to be interpreted through existing MAS 
failure categories; details in 
Appendix~\ref{app:more_profile}.

\textbf{MAS Execution Setup. }
Following the task instantiation in
Figure~\ref{fig:puzzle}, we utilize two core \textsc{Solve\_Clue} domains, \textit{Math} and \textit{Code}. For each domain, we generate 100 rooms, with 25 rooms at each difficulty level, resulting in 200 unique rooms. We execute MAS using five popular LLM backbones: Claude Sonnet 4.6~\citep{anthropic2025system}, Claude Opus 4.8~\citep{anthropic2026opus48}, DeepSeek-R1~\citep{guo2025deepseek}, GPT-4.1~\citep{openai2025gpt41} and GPT-5.4~\citep{openai2026gpt54}. For each backbone, we run both \textit{Tool} and \textit{No-Tool} modes, with external tool use enabled or disabled, respectively. LLMs without temperature control (e.g., Claude Opus 4.8) are run with the default settings, while the remaining LLMs run with temperatures of 0.0, 0.3, and 0.6. Each trace is automatically annotated with ground-truth labels at the room, puzzle, and action levels. 
Trace token costs are reported in Table~\ref{tab:api-cost} in the Appendix.

\begin{table}[ht]
\centering
\caption{Room-Level and Puzzle-Level MAS Success Rate ($\uparrow$) Across LLM Types on \appfull{}. \fcolorbox{hlblue}{hlblue}{\textbf{Best}} and \colorbox{hlorange}{worst} values are highlighted. All rooms are fixed across all LLM instantiations
to ensure fair comparison. Success Rate (SR) is verified by the final Room/Puzzle output.}
\vspace{-8pt}
\label{fig:room_sr}
\resizebox{\textwidth}{!}{%
\begin{tabular}{lc cccc c cc}
\toprule
\multirow{2}{*}{\textbf{LLM}} & \multirow{2}{*}{\textbf{\# Room ($\times$Temp)}} & \multicolumn{5}{c}{\textbf{Room-Level SR}} & \multicolumn{2}{c}{\textbf{Puzzle-Level}} \\
\cmidrule(lr){3-7} \cmidrule(lr){8-9}
 & & \textbf{Easy} & \textbf{Medium} & \textbf{Hard} & \textbf{Extreme} & \textbf{Overall} & \textbf{Attempted / Total} & \textbf{SR} \\
\midrule
Claude Opus 4.8 & 400 ($\times$1) & \colorbox{hlblue}{\textbf{70.0\%}} & \colorbox{hlblue}{\textbf{63.0\%}} & \colorbox{hlblue}{\textbf{29.0\%}} & \colorbox{hlblue}{\textbf{30.0\%}} & \colorbox{hlblue}{\textbf{48.0\%}} & 805 / 1000 & \colorbox{hlblue}{\textbf{61.0\%}} \\
Claude Sonnet 4.6 & 400 ($\times$3) & 52.3\% & 35.3\% & 11.0\% & 19.3\% & 29.5\% & 2065 / 3000 & 41.4\% \\
DeepSeek-R1 & 400 ($\times$3) & 66.7\% & 42.0\% & 12.3\% & 22.0\% & 35.8\% & 2054 / 3000 & 43.9\% \\
GPT-4.1 & 400 ($\times$3) & \colorbox{hlorange}{46.3\%} & \colorbox{hlorange}{19.0\%} & \colorbox{hlorange}{3.0\%} & \colorbox{hlorange}{0.0\%} & \colorbox{hlorange}{17.1\%} & 1521 / 3000 & \colorbox{hlorange}{21.0\%} \\
GPT-5.4 & 400 ($\times$3) & 60.0\% & 27.0\% & 10.3\% & 0.3\% & 24.4\% & 1689 / 3000 & 30.0\% \\
\bottomrule
\end{tabular}%
}
\end{table}

\textbf{\app{} is Challenging for SOTA LLMs. }
We report the overall success rate of the MAS execution on the room and puzzle level in Table~\ref{fig:room_sr}. 
At both levels, performance drops sharply as the overall difficulty increases. Claude Opus 4.8 ranks first on both levels, while GPT-4.1 ranks last across both levels. 

\begin{table}[ht]
\centering
\caption{Success Rate ($\uparrow$) of Actions on \appfull{}. For every action, we annotate the primary LLM capability it evaluates.}
\vspace{-8pt}
\label{tab:action-success-rate}
\resizebox{\textwidth}{!}{%
\begin{tabular}{l c c c c c c c c}
\toprule
\multirow{2}{*}{\textbf{LLM}} & \textbf{SEL.\_PUZZLE} & \textbf{OBS.\_CLUE} & \textbf{SOL.\_CLUE} & \textbf{OBS.\_INSTRU} & \textbf{TOOL\_CALL} & \textbf{APL.\_DELTA} & \textbf{OBS.\_PUZZLE} & \textbf{Overall} \\
 & \textit{(Planning)} & \textit{(Distraction)} & \textit{(Math)} & \textit{(Distraction)} & \textit{(Tool Use)} & \textit{(Reasoning)} & \textit{(Verification)} &  \\
\midrule
Claude Opus 4.8 & 94.8\% & \colorbox{hlblue}{\textbf{99.8\%}} & 80.4\% & \colorbox{hlblue}{\textbf{97.8\%}} & \colorbox{hlblue}{\textbf{73.1\%}} & \colorbox{hlblue}{\textbf{75.8\%}} & 98.9\% & \colorbox{hlblue}{\textbf{89.1\%}} \\
Claude Sonnet 4.6 & 97.1\% & 97.5\% & \colorbox{hlblue}{\textbf{80.5\%}} & 95.4\% & 64.9\% & 60.1\% & \colorbox{hlblue}{\textbf{100.0\%}} & 85.0\% \\
DeepSeek-R1 & \colorbox{hlblue}{\textbf{100.0\%}} & \colorbox{hlorange}{46.2\%} & \colorbox{hlorange}{63.0\%} & \colorbox{hlorange}{73.3\%} & \colorbox{hlorange}{60.8\%} & 64.1\% & \colorbox{hlorange}{97.0\%} & \colorbox{hlorange}{68.4\%} \\
GPT-4.1 & 70.9\% & 97.2\% & 71.3\% & 94.3\% & 63.2\% & \colorbox{hlorange}{41.5\%} & 98.4\% & 78.3\% \\
GPT-5.4 & \colorbox{hlorange}{65.6\%} & 98.3\% & 78.7\% & 93.5\% & 67.0\% & 53.3\% & 99.9\% & 82.2\% \\
\midrule
\textbf{Average} & 85.7\% & 87.8\% & 74.8\% & 90.9\% & 65.8\% & 59.0\% & 98.8\% & 80.6\% \\
\bottomrule
\end{tabular}%
}
    \vspace{-10pt}
\end{table}


\begin{wrapfigure}{l}{0.45\textwidth}
    \centering
    \vspace{-10pt}
    \includegraphics[width=\linewidth]{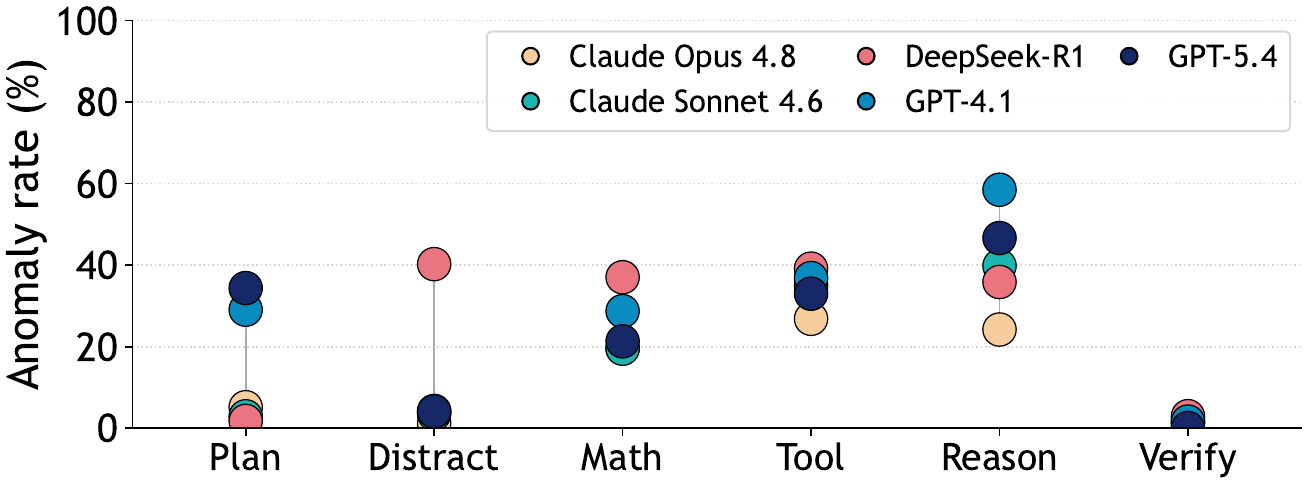}
    \vspace{-12pt}
    \caption{Anomaly rate by capability $\times$ LLMs. }
    \vspace{-8pt}
    \label{fig:llm_lollipop}
\end{wrapfigure}

\textbf{Anomalies Across Capabilities and LLM Backbones.}
We 
characterize anomalies at the finest granularity, \ie, the action level. As shown in Table~\ref{tab:action-success-rate} and Figure~\ref{fig:llm_lollipop}, anomaly patterns are strongly \emph{capability-specific} and vary substantially across LLM backbones. Verification action
is consistently reliable, with anomaly rates below $3\%$, whereas tool use remains challenging across models ($\sim27$--$39\%$). Reasoning shows even larger variation, ranging from roughly $24\%$ to $59\%$, indicating substantial backbone-specific differences. Planning and distraction handling are similarly model-dependent, with anomaly rates varying from near $0\%$ (zero) to around $40\%$ across backbones.
It is interesting to observe that each backbone exhibits a distinct anomaly pattern over different capabilities.




\begin{tcolorbox}[colback=gray!8,colframe=black!15,boxrule=0.6pt,arc=2pt,left=4pt,right=4pt,top=1pt,bottom=1pt]
\small\textbf{Observation 1 [Backbone Effect].}
\textit{Stronger LLMs do not uniformly reduce anomalies; instead, each backbone exhibits distinct capability and anomaly profiles.}
\end{tcolorbox}

\textbf{Silent vs. Loud Anomalies.}
We divide failures into \emph{silent}, \ie failed actions within solved puzzles, and \emph{loud} anomalies, \ie failed actions within failed puzzles. 
Surprisingly, 
Table~\ref{tab:silent-failures} shows that overall, $7.7\%$--$30.4\%$ anomalies are silent. By action, planning errors 
(\textsc{Select\_Puzzle}) are always silent, the reason may be that
they affect 
long-term planning rather than current puzzle outcomes; final-step errors (\textsc{Apply\_Delta}) are always loud, as incorrect updates directly lead to puzzle failure; and intermediate 
actions (\textsc{Solve\_Clue}, \textsc{Observe\_Instrument}, \textsc{Tool\_Call}) 
show mixed behavior. Both types matter for MAS monitoring: 
\textit{loud anomalies are task-fatal, while silent anomalies reveal latent behavioral deviations.}


\begin{tcolorbox}[colback=gray!8,colframe=black!15,boxrule=0.6pt,arc=2pt,left=4pt,right=4pt,top=1pt,bottom=1pt]
\small\textbf{Observation 2 [Observability Gap].} \textit{MAS anomalies exhibit different levels of observability, requiring AD methods to remain effective across both explicit task failures and subtler intermediate deviations.}

\end{tcolorbox}

\begin{table}[ht]
\centering
\caption{Silent failure rate by model and failure type on \appfull{}. Formatting errors (\textbf{Fmt Err}), \eg invalid JSON outputs, separated from action failures. Each column reports fraction of silent failures among all failures of  corresponding type. ``--'' indicates no such failure occurred.}
    \vspace{-6pt}
\label{tab:silent-failures}
\resizebox{\textwidth}{!}{%
\begin{tabular}{l c c c c c c c c c}
\toprule
\textbf{LLM} & \textbf{Overall} & \textbf{Fmt Err} & \textbf{SEL.\_PUZZLE} & \textbf{OBS.\_CLUE} & \textbf{SOL.\_CLUE} & \textbf{OBS.\_INSTRU} & \textbf{TOOL\_CALL} & \textbf{APL.\_DELTA} & \textbf{OBS.\_PUZZLE} \\
\midrule
Claude Opus 4.8 & 9.0\% & 9.1\% & 100.0\% & 0.0\% & 10.2\% & 16.7\% & 14.5\% & 0.0\% & 0.0\% \\
Claude Sonnet 4.6 & 7.7\% & 18.8\% & 100.0\% & 0.0\% & 7.9\% & 42.1\% & 10.5\% & 0.0\% & -- \\
DeepSeek-R1 & 30.4\% & 34.1\% & -- & -- & 1.5\% & 50.0\% & 7.5\% & 0.0\% & 100.0\% \\
GPT-4.1 & 12.2\% & 10.9\% & 100.0\% & 13.9\% & 13.5\% & 10.8\% & 25.5\% & 0.0\% & 80.0\% \\
GPT-5.4 & 11.4\% & 18.5\% & 100.0\% & 10.3\% & 4.2\% & 20.0\% & 2.5\% & 0.0\% & 0.0\% \\
\bottomrule
\end{tabular}%
}
\end{table}


\begin{figure}[t]
    \centering
\includegraphics[width=0.99\linewidth]{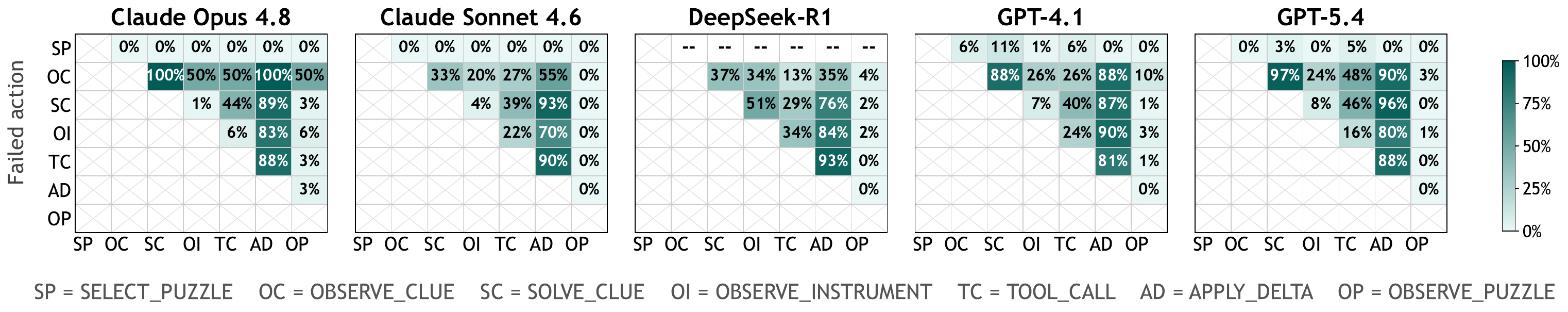}
\vspace{-5pt}
\caption{Action-level anomaly propagation by LLMs. \textbf{Row}: failed source action, \textbf{Column}:  subsequent failed actions.  \textbf{Colored cells}: propagation rates, as fraction of source failures followed by a downstream failure of  target action. Darker colors indicate stronger propagation. \textbf{``--'' cells}: no source failures. \textbf{Cross-hatched cells}: invalid action pairs. }

\label{fig:propagation}
\vspace{-15pt}
\end{figure}

\textbf{Anomaly Propagation.}
Figure~\ref{fig:propagation} shows that anomalies generally propagate along the
workflow order. Across GPT and Claude backbones, early perception failures
(\textsc{Observe\_Clue}) often lead to downstream failures in
\textsc{Solve\_Clue}, \textsc{Tool\_Call}, and especially
\textsc{Apply\_Delta}.
This indicates that incorrect upstream observations can
contaminate both reasoning and final execution. \textsc{Apply\_Delta} acts as a
common sink for propagated errors, since failures in solving, instrument
observation, or tool use often surface as incorrect state updates. DeepSeek-R1
shows a less concentrated pattern, with failures spread more broadly across
observation, solving, and tool-use steps. Overall, propagation is shaped both by
the task workflow and by backbone-specific failure profiles.

\begin{tcolorbox}
[colback=gray!8,colframe=black!15,boxrule=0.6pt,arc=2pt,left=4pt,right=4pt,top=1pt,bottom=1pt]
\small\textbf{Observation 3  [Heterogeneous Propagation].} \textit{Anomaly propagation patterns are shaped by both the causal structure of the workflow and the LLM backbone. 
}
\end{tcolorbox}

\section{Evaluating
Anomaly Detection Strategies on \app{}}
\label{sec:exp}


We now use \app{} traces and labels to evaluate AD methods. We select a core subset \appeval{} from \appfull{} with temperature $=0$, as they yield more stable model behavior for AD evaluation. Our evaluation assesses the
effectiveness of AD methods on MAS anomalies and characterizes MAS anomaly detectability to inform future research into MAS-tailored detectors. 
Specifically, we conduct experiments to answer
\textbf{RQ1}(\S\ref{sec: main_result}): Which classes of AD methods are 
effective for MAS anomaly detection?
\textbf{RQ2} (\S\ref{sec:detectability}): Which MAS anomalies are most detectable by which methods?
\textbf{RQ3} (\S\ref{sec:propagation}): How 
early in the long trace can 
AD methods capture anomaly propagation? 
Below,
each subsection addresses
one of the RQs; including
analysis of 
results, key takeaways, and proposal of future MAS AD research. 

\subsection{Experimental Setup}
\label{sec: setup}

\textbf{Dataset. }\appeval{} contains 17,180 labeled actions, with an anomaly ratio of 19.68\%.
We split the dataset into training and testing sets
using an 8:2 stratified split.
To support different AD 
families,
we construct method-specific
inputs, including tabular features such as token counts and
embeddings, graph representations, MAS-aware interaction graphs,
and raw-text traces for LLM-based detectors.
Full data pre-processing details are provided in
Appendix~\ref{app:data_processing}.


\textbf{Methods and Evaluation Metrics.}
We evaluate 25 AD methods across {\it four supervision settings: }
unsupervised, one-class classification (OCC), semi-supervised 
with 5\% labeled anomalies following existing benchmarks~\citep{ADBench}, and fully supervised. These methods span four families: tabular, graph-based, MAS-specific, and LLM-as-detector methods. Representative baselines include XGBoost~\citep{chen2016xgboost}, Random Forest~\citep{breiman2001random}, DevNet~\citep{pang2019deep}, DeepSAD~\citep{ruff2019deep}, OC-SVM~\citep{scholkopf2001estimating}, DOMINANT~\citep{ding2019deep}, TAM~\citep{qiao2023truncated}, GGAD~\citep{qiao2024generative}, G-Safeguard~\citep{gsafeguard}, BlindGuard~\citep{BlindGuard}, and Gemma4~\citep{gemma4_deepmind_2026}; 
with full list of methods reported in Figure \ref{fig:overall_aucroc}, and the method details
and computation resources in Appendix~\ref{app:methods}. We 
report AUC-ROC here, with F1-score, accuracy, and balanced accuracy in the Appendix \ref{app:more_results}. 

\subsection{Overall Effectiveness of Existing AD Methods}

\label{sec: main_result}

\begin{figure}[t]
\centering

    \includegraphics[width=0.9\linewidth]{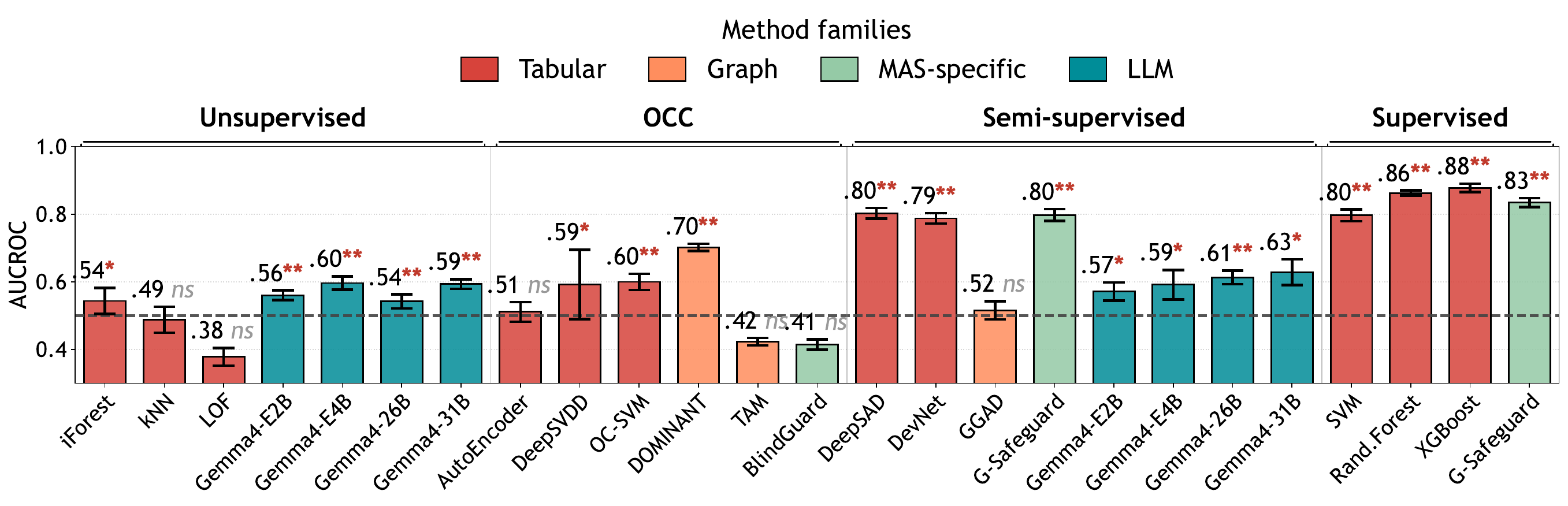}
    \vspace{-10pt}
\caption{Overall AUCROC of AD methods categorized by supervision level and method family. LLM zero-shot methods are grouped as unsupervised, while LLM few-shot methods are grouped as semi-supervised. \textbf{Dashed line}: A random baseline with AUCROC=0.5 as reference. \textbf{Significance vs. Random}: ** $p<0.01$, * $p<0.05$, ns $p \geq0.05$.  More metrics with statistical tests are reported in Table \ref{tab:main_results} and  \ref{tab:main_results_pvalues_n5} in  Appendix~\ref{app:more_results}. }
\label{fig:overall_aucroc}
\vspace{-12pt}
\end{figure}

\textbf{Detection Performance Across Supervision Levels.}
Figure~\ref{fig:overall_aucroc} reports each method's mean AUCROC. Method performance generally increases as stronger anomaly supervision becomes available. LLM-based unsupervised detectors and the strongest graph-based OCC methods extract more meaningful signal. Supervision (Semi- and fully-supervised) yields substantial gains, with top methods (XGBoost, G-Safeguard, Random Forest) reaching AUROC $\geq 0.80$. Notably, the supervision gap on \app{} ($\sim$$0.28$ AUROC) is substantially larger than on tabular ADBench ($\sim$$0.20$ between best unsupervised and fully-supervised methods)~\citep{ADBench}. This suggests that MAS AD is challenging for existing methods without explicit anomaly signals. However, large-scale step-level anomaly labels are often expensive to obtain in deployed MAS.

\begin{tcolorbox}[colback=gray!8,colframe=black!15,boxrule=0.6pt,arc=2pt,left=4pt,right=4pt,top=1pt,bottom=1pt]
\small\textbf{Takeaway 1: MAS Anomalies Are Particularly Difficult to Detect Without Supervision. } Compared with traditional AD, MAS AD exhibits a substantially larger supervision gap, suggesting that MAS-specific anomaly signals are harder to discover without labels.
\end{tcolorbox}



\textbf{Detection Performance Across Method Families.} We compare method families within each supervision level
(Figure~\ref{fig:overall_aucroc}). The winning family shifts across
supervision. As described before, only LLM zero-shot extracts meaningful
signal (Gemma $0.54$--$0.60$) in the unsupervised setting. However, performance shows no clear gain with model size. In contrast, few-shot supervision yields only modest overall improvement, but introduces a clearer scaling benefit, with larger Gemma variants performing better. This suggests that model scale becomes more useful when limited anomaly supervision is available. In OCC, graph-based DOMINANT performs best ($0.70$), while TAM remains near the random baseline ($0.42$), indicating that graph-based modeling is not uniformly beneficial for MAS traces. With semi- or full supervision, MAS-specific and tabular methods dominate: DeepSAD and G-Safeguard tie atop semi-supervised ($0.80$), while XGBoost leads supervised ($0.88$), followed by Random Forest ($0.86$) and G-Safeguard ($0.83$).

\begin{tcolorbox}[colback=gray!8,colframe=black!15,boxrule=0.6pt,arc=2pt,left=4pt,right=4pt,top=1pt,bottom=1pt]
\small\textbf{Takeaway 2: Designing for MAS Matters.} 
The most effective methods exploit MAS-specific signals, \ie semantic 
understanding (LLMs with lower supervision) or MAS-specific agent interaction 
modeling. 
\end{tcolorbox}

\textit{\textbf{Future Directions: Toward Low-cost and Label-efficient MAS AD.}}
Our takeaways answer \textbf{RQ1}: existing AD methods are most effective for MAS AD only under strong supervision or are tailored to MAS-specific signals. Future MAS AD research should aim to close this gap with less-supervised and MAS-tailored methods. While LLMs offer strong semantic understanding of the abnormal traces, our cost analysis (Figure \ref{fig:cost_auc}, Appendix~\ref{app:methods}) shows a 2-order-of-magnitude cost gap between LLM-based and conventional detectors, making LLM-as-a-judge costly for long MAS traces. This motivates cost-effective LLM-based detection through model cascading~\citep{nie2024online}, distillation~\citep{hinton2015distilling}, or selective invocation~\citep{ding2024hybrid}. For non-LLM or hybrid directions, promising directions include better modeling of MAS semantics and agent interactions.


\subsection{Detection Performance Across Different MAS Anomaly Types}
\label{sec:detectability}

We now study anomaly detectability across AD methods. We categorize MAS anomalies based on our profiling in \S\ref{sec:execution}, based on: (1) the action and LLM backbone in which they arise (Observation~1), and (2) their observability as silent or loud anomalies (Observation 2).

\begin{figure}[t]
    \centering
    \includegraphics[width=\linewidth]{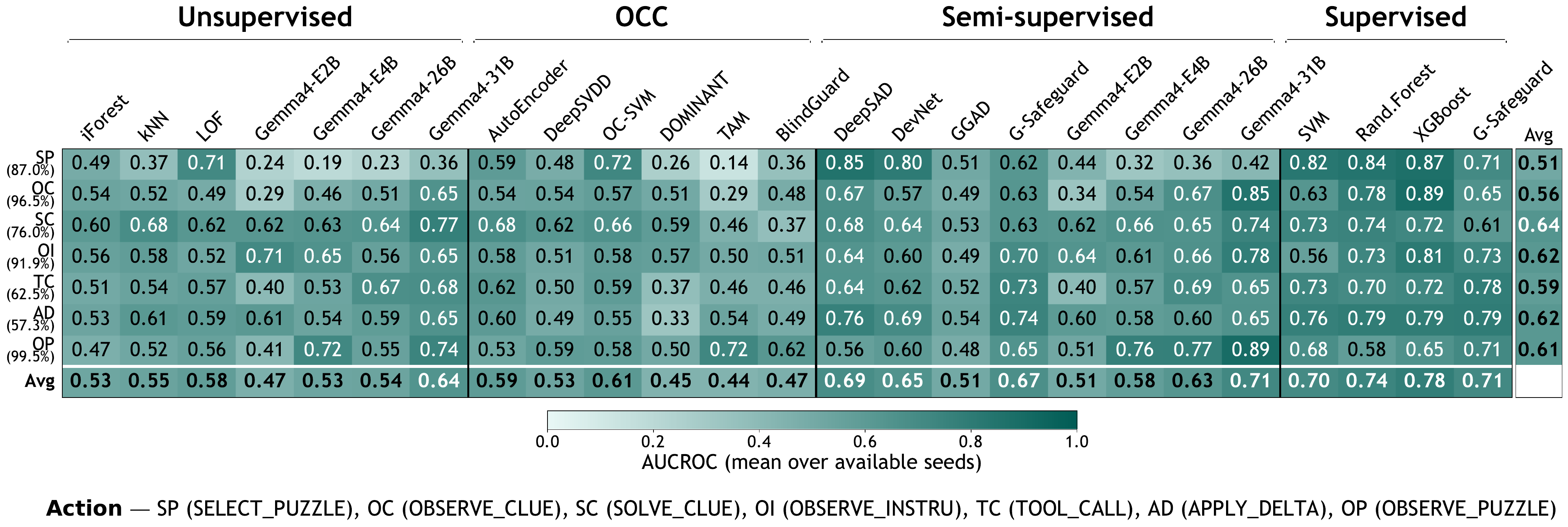}
    \vspace{-12pt}
\caption{
AUCROC of AD methods by actions (full results with std in Table \ref{tab:aucroc-by-action} Appendix~\ref{app:more_results}). \textbf{Rows:} Six action types with  success rate; \textbf{Bottom row:} avg AUCROC per method across actions. \textbf{Columns:} Method AUC. \textbf{Rightmost column:} avg AUCROC per action across methods. 
}
    \label{fig:step-eval}
\end{figure}

\textbf{Anomaly Detectability vs. Action Success.}
Figure~\ref{fig:step-eval} reports each action's success rate and AUCROC across methods. Action success and detectability correlate only weakly: \textsc{Solve\_Clue} achieves the highest average AUCROC ($0.64$) despite a moderate success rate ($76.0\%$), while \textsc{Observe\_Puzzle} has the highest success rate ($99.5\%$) but slightly lower detectability ($0.61$). Notably, \textsc{Select\_Puzzle}, which requires long-horizon reasoning over cross-puzzle dependencies, is the hardest to detect (avg AUCROC $0.51$) despite a relatively high success rate ($87.0\%$). This suggests that anomalies involving long-horizon reasoning may be particularly challenging for AD methods.

\begin{tcolorbox}[colback=gray!8,colframe=black!15,boxrule=0.6pt,arc=2pt,left=4pt,right=4pt,top=1pt,bottom=1pt]
\small\textbf{Takeaway 3: Existing AD Methods Struggle to Detect Anomalies in Long-Horizon Reasoning.}
Actions requiring long-horizon reasoning exhibit the lowest detectability, highlighting a key challenge for current AD methods.
\end{tcolorbox}



\begin{figure}[t]

    \centering
    \includegraphics[width=\linewidth]{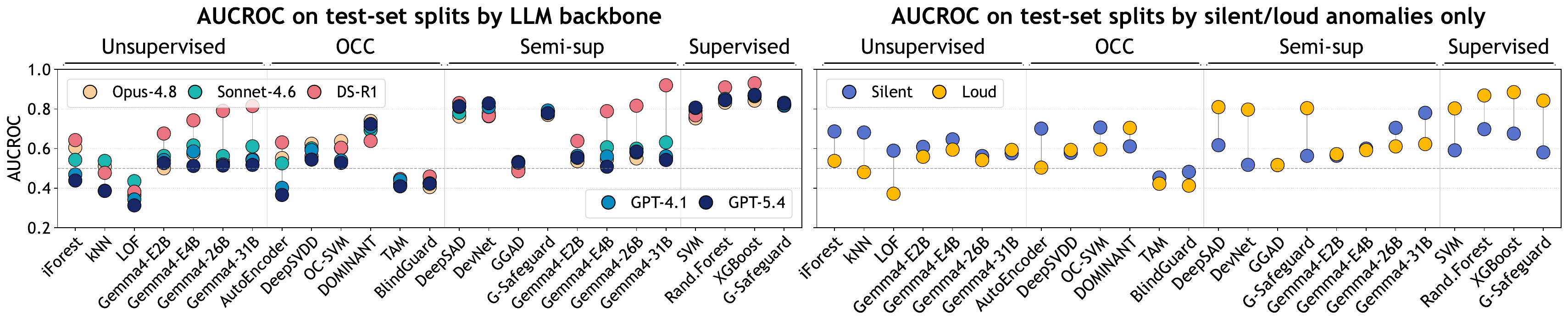}
    \vspace{-16pt}
    \caption{AD performance for different anomaly types. Evaluated \textbf{Left}: on test sets split by LLM backbone. 
\textbf{Right}: on test sets split by anomaly type (silent: puzzle solved; 
loud: puzzle failed).}
    \label{fig:method_vs_anomaly}
    \vspace{-16pt}
\end{figure}

\textbf{Anomaly Detectability vs.\ LLM Backbones.}
Figure~\ref{fig:method_vs_anomaly} (left) reports detection performance on test sets split by LLM backbone. Detection is generally easier on DeepSeek-R1 traces and harder on GPT-series traces. Backbone sensitivity is most pronounced for LLM-based detectors (Gemma variants), with AUCROC gaps approaching $0.38$ (Gemma-4-31B few-shot) across backbones.  In contrast, semi- and fully-supervised methods are substantially more stable across backbones, suggesting that anomaly supervision improves robustness to backbone-specific anomaly patterns.

\begin{tcolorbox}[colback=gray!8,colframe=black!15,boxrule=0.6pt,arc=2pt,left=4pt,right=4pt,top=1pt,bottom=1pt]
\small\textbf{Takeaway 4: Method Robustness Varies Across LLM Backbones.}  Effective evaluation of MAS AD therefore requires testing 
across multiple LLM backbones.
\end{tcolorbox}

\textbf{Anomaly Detectability vs.\ Observability.} We compare the detectability of silent and loud anomalies, as defined in Table~\ref{tab:silent-failures}. Since silent format errors are easy to detect, we focus on the remaining harder silent anomalies. Figure~\ref{fig:method_vs_anomaly}(right) reveals a supervision-dependent observability gap. Under unsupervised and OCC settings, loud anomalies are often harder to detect than silent ones, indicating that task-fatal failures do not necessarily produce stronger local anomaly signals. With stronger non-LLM supervision, this trend reverses: semi- and fully-supervised detectors detect loud anomalies more reliably, with supervised methods reaching $0.80$--$0.88$ AUCROC. In contrast, the performance of LLM detectors is much more robust to the observability of the anomalies. 

\begin{tcolorbox}[colback=gray!8,colframe=black!15,boxrule=0.6pt,arc=2pt,left=4pt,right=4pt,top=1pt,bottom=1pt]
\small\textbf{Takeaway 5: Loud Anomalies Require Supervision.}
Loud anomalies can look locally plausible enough to evade detection; supervision primarily improves detection of these task-fatal failures. 
\end{tcolorbox}


\textit{\textbf{Future Directions: Beyond Surface Anomaly Detection.}}
Our takeaways answer \textbf{RQ2}: the hardest anomalies are logical, task-failing, and produced by well-formatted backbones. Although supervision and MAS-specific design help, these findings motivate three directions. First, logical reasoning failures call for richer signals beyond surface embeddings, including LLM-internal features such as hidden states~\citep{hidden_state_forensics,abdelnabi2025get, xu2026when}, attention patterns~\citep{he2025attention}, and token-level uncertainty~\citep{semantic_entropy_probes,icr_probe}. Second, task-failing anomalies may remain locally plausible, motivating consistency- or verification-based detectors such as inter-agent checks and debate~\citep{march_multi_agent,markov_chain_debate}. Third, backbone-induced variance up to $0.30$ AUROC motivates backbone-robust detectors and multi-backbone evaluation enabled by \app{}~\citep{irad_anomaly_detection}.

\subsection{Detection Performance under MAS Anomaly Propagation}
\label{sec:propagation}

\begin{wrapfigure}{l}{0.4\textwidth}
    \centering
    \vspace{-10pt}
    \includegraphics[width=\linewidth]{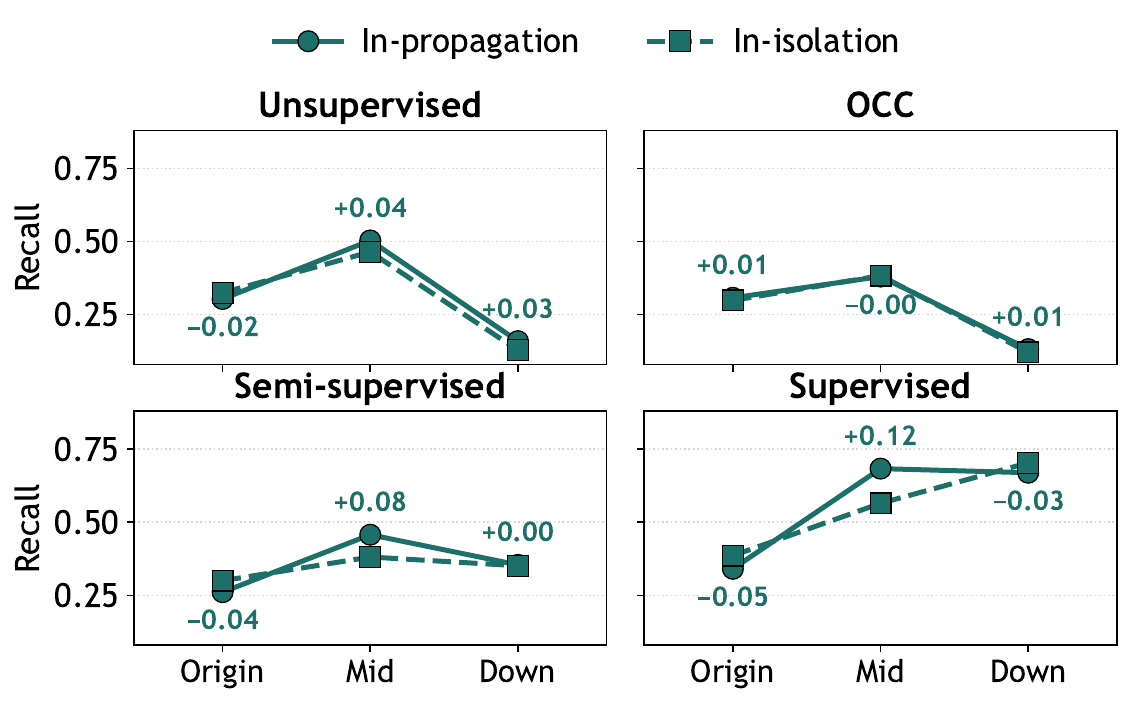}
    \vspace{-10pt}
    \caption{Recall on anomaly propagation chains. 
    \textbf{In-propagation (solid)}: recall on an action inside 
    propagation chains; \textbf{In-isolation (dashed)}: global 
    recall on the same actions as reference. The
     delta of the two is annotated. $-$ means the anomalies are \textit{harder} to detect inside a propagation chain, $+$ otherwise. }
    \label{fig:detect_propagation}
    \vspace{-10pt}
\end{wrapfigure}

To analyze detection under anomaly propagation, we identify propagation chains as puzzles with multiple failed actions. We label each failed action by its chain role: \textbf{Origin} (first), \textbf{Mid}dle (intermediate), or \textbf{Down}stream (last). We report the average recall of each supervision family across these roles in Figure~\ref{fig:detect_propagation}, since identifying propagation early is the focus. 

Across supervision families, propagation has a role-dependent effect on anomaly
detectability. Origin steps generally stay harder in propagation, with 
lower recall (delta down to $-0.04$) on origins than on isolated failures. The strongest propagation
effect appears at middle steps, where recall increases for most families and
reaches the largest gain under supervised methods ($+0.12$). Downstream effects
are more mixed: unsupervised, OCC, and semi-supervised detectors show small
gains, whereas supervised detectors perform slightly worse than their isolated
reference. These results suggest that propagation most clearly exposes
intermediate accumulated errors. A detailed per-method, per-role
breakdown is provided in Figure~\ref{fig:propagate_deteailed} of
Appendix~\ref{app:more_results}.

\begin{tcolorbox}[colback=gray!8,colframe=black!15,boxrule=0.6pt,arc=2pt,left=4pt,right=4pt,top=1pt,bottom=1pt]
\small\textbf{Takeaway 6: Root Causes Remain Hard to Detect.}
Propagation more consistently exposes accumulated middle-step anomalies, while root-cause failures often remain subtle and detector-dependent.
\end{tcolorbox}

\textit{\textbf{Future Directions: Detecting Subtle Root-Cause 
Anomalies.}}
Our takeaways answer \textbf{RQ3}: in propagation chains, AD 
methods detect downstream anomalies more easily than in 
isolation but miss origin failures more often; stronger supervision widens this asymmetry. Two directions follow. First, AD should 
move beyond isolated step features to propagation-aware models 
that exploit inter-step dependencies, surfacing anomalies whose 
signals span multiple steps.  Insights from causality-aware AD in 
related domains~\citep{causality_invariant_ad,carots_causal} can 
inform such designs. Related insights from process-reward modeling show that step-level progress signals improve credit assignment over outcome-only supervision~\citep{setlur2025rewarding}, motivating similar step-attribution mechanisms for MAS AD. Second, \textit{root-cause localization}
goes beyond detection by 
identifying which step initiates a cascade. Existing 
failure-attribution works for LLM 
agents~\citep{agenttrace_root_cause,whoandWhen,AgenTracer} 
address this post-hoc on completed traces, while real-time MAS 
AD requires proactive localization during execution. 
\section{Conclusion}
\label{sec:conclusion}

MAS anomaly detection needs a refreshable benchmarking paradigm to remain useful as LLMs evolve. We propose the innovative \app{}
paradigm that tackles the key challenges, instantiate \appfull{} as a labeled-trace dataset, 
and evaluate diverse 
SOTA AD methods on it. 
Results show MAS AD remains largely unsolved, motivating the need for the development of AD detectors tailored to multi-agent settings. 
As a key contribution, our \app{} design principles, namely, generative tasks, reproducible traces, and deterministic step-level labels, 
future-proof MAS AD benchmarking and method development amid rapid LLM evolution.


\textbf{Limitations.} Although \app{} covers diverse capabilities and domains, we acknowledge
it does not fully exhaust the
complete
 MAS design space, where variations in agent organization, communication, and
memory strategies
may introduce new anomaly surfaces and propagation patterns. A full discussion can be found in Appendix \ref{app:discussion}.

\textbf{Broader Impact.} \app{} provides an adaptive testbed for developing and evaluating future MAS anomaly detection methods. By enabling systematic MAS AD research on rapidly evolving LLMs and agent systems, MAADBench contributes to the broader goal of building reliable LLM-based multi-agent systems and thus
furthers  their future application for solving real-world problems.

\pdfbookmark[1]{References}{References}
\bibliography{references}
\bibliographystyle{preprint}

\newpage
\appendix
\section{Related Work}
\label{app:related_work}

\noindent\textbf{MAS Failure Analysis and Robustness.}
A growing body of work studies MAS reliability~\citep{guo2024large} through two 
complementary lenses. The first focuses on failure analysis: 
collecting execution traces from deployed systems and annotating 
them to understand why failures occur~\citep{MAST, Trail, 
whoandWhen, AgenTracer}. MAST~\citep{MAST} establishes the most 
comprehensive failure taxonomy to date, covering 14 failure modes 
across 1,600+ traces from 7 frameworks. TRAIL~\citep{Trail} focuses 
on error categorization and span-level localization within agentic 
traces. Who\&When~\citep{whoandWhen} further attributes failures to 
specific agents and decisive error steps. AgenTracer~\citep{AgenTracer} 
automates trace annotation through counterfactual replay and 
programmatic fault injection. MP-Bench~\citep{MP-bench} revisits 
failure attribution from a multi-perspective standpoint, showing 
that MAS failures often admit multiple plausible attributions due 
to complex inter-agent dependencies and ambiguous execution 
trajectories. Silent Failures~\citep{SilentFailures} is closest in 
spirit to MAS AD, curating two labeled datasets for anomaly 
detection, but the datasets have not been publicly released. 
The second lens focuses on robustness and security, including 
defenses against adversarial perturbations and agent 
hijacking~\citep{gsafeguard, BlindGuard, yang2024watch}. While 
both lines of work provide valuable insights, they focus on 
post-hoc analysis or targeted defense rather than providing 
long-lived benchmark infrastructure for systematic MAS AD 
evaluation. 

\noindent\textbf{Anomaly Detection Benchmarks.}
In the broader AD literature, benchmark design has matured 
significantly across multiple data modalities. ADBench~\citep{ADBench} 
provides the most comprehensive tabular AD benchmark, examining 
30 algorithms across 57 datasets along three angles: supervision 
level, anomaly type, and robustness to data corruption. 
TSB-AD~\citep{TSBAD} addresses systematic flaws in time-series AD 
benchmarking, including dataset quality, evaluation metric bias, 
and hyperparameter tuning. GADBench~\citep{GAD-bench} benchmarks 
supervised graph anomaly detection, comparing 29 models across 
10 real-world datasets and finding that simple tree ensembles 
with neighborhood aggregation can outperform specialized GNNs. Complementing GADBench's supervised graph AD arena, UB-GOLD \citep{wang2025unifying} unifies unsupervised graph-level AD and OOD detection under one benchmark, exemplifying the shift toward unified evaluation across supervision regimes. 
Together, these works demonstrate the importance of structured 
benchmark angles, accurate labeling, and careful dataset curation 
for drawing meaningful conclusions from AD experiments. \app{} is 
inspired by this design philosophy but addresses challenges unique 
to the MAS setting: rather than curating existing datasets, we 
generate traces from a live MAS framework, and rather than relying 
on human or algorithmic annotation, we obtain step-level labels 
through deterministic oracle verification.

\noindent\textbf{LLM Agent Evaluation and Step-level Supervision.}
A large body of work constructs task environments and benchmarks 
for evaluating LLM agent capabilities across diverse domains. 
AgentBench~\citep{agentbench} is the first systematic 
multi-environment benchmark for LLM-as-agent, covering 8 
environments including OS, database, web browsing, and games. 
SWE-bench~\citep{SWE} evaluates agents on real GitHub issues; 
WebArena~\citep{zhou2024webarena} and GAIA~\citep{GAIA} test 
open-ended web navigation and general assistant capabilities; 
OSWorld~\citep{OSWorld} benchmarks multimodal agents on 
real computer environments; AppWorld~\citep{AppWorld} 
tests interactive coding agents in a controllable world of apps; 
and TheAgentCompany~\citep{xu2024theagentcompany} evaluates agents 
on consequential real-world professional tasks. Comprehensive 
surveys on LLM agent evaluation further systematize these 
efforts~\citep{mohammadi2025evaluation}. 
Despite their breadth, these benchmarks and surveys focus 
primarily on task completion rather than behavioral anomaly 
detection, and their verification mechanisms:  unit tests, 
human judgment, or execution-based scripts,  do not scale well to 
step-level labeling of agent traces. A complementary line of 
work has demonstrated the value of fine-grained process 
supervision: PRM800K~\citep{prm800k} and 
Math-Shepherd~\citep{math_shepherd} show that step-level 
labels provide substantially better learning signal than the sparse 
outcome-level supervision alone, motivating the need for 
step-level ground truth in \app{}. Concurrent work 
MM-Escape~\citep{mmEscape} independently adopts escape rooms as 
an LLM evaluation environment, focusing on multimodal perception 
in a single-agent setting. In contrast, \app{} instantiates 
escape rooms specifically for MAS anomaly detection, with 
multi-agent collaboration, generative task construction, and 
deterministic oracle-based step-level labeling as first-class 
design principles.

\noindent\textbf{Benchmark Contamination and Long-lived Benchmarks.}
Benchmark contamination--- test data leaks into LLM 
training corpora and inflates evaluation scores, has become 
a central concern in LLM evaluation~\citep{magar2022contamination, 
golchin2024timetravelllms, deng2024survey, deng2023investigating, 
xu2024benchmarking}. The field has responded by shifting from 
static to dynamic benchmarking paradigms~\citep{chen2025staticdynamic}. 
One line of work mitigates contamination by continuously sourcing 
questions from recent information: LiveBench~\citep{white2025livebench} 
releases new questions monthly from recent math competitions, 
arXiv papers, and news articles, while 
LiveCodeBench~\citep{jain2025livecodebench} continuously collects 
new coding problems from online platforms. A complementary 
approach uses symbolic templates and parameter sampling to 
generate novel instances on demand: 
GSM-Symbolic~\citep{GSM-Symbolic} generates diverse math questions 
from symbolic templates, demonstrating that LLM performance on 
fixed benchmarks may reflect memorization rather than genuine 
reasoning. \app{} adopts the template-based paradigm but extends 
it to the MAS AD setting: rather than sampling numerical 
parameters into fixed question templates, we compose entire task 
instances from parameterizable puzzle subtasks with hidden 
dependency structure, yielding a large combinatorial instance 
space that makes contamination practically infeasible.

\section{Discussion and Limitations}
\label{app:discussion}

\textbf{Benchmark Refresh and Anomaly Drift.}
A long-lived benchmark should not assume that anomaly distributions remain stationary as MAS components evolve. New LLM backbones, decoding configurations, or system components may induce different failure patterns, causing detectors trained on earlier traces to degrade. The value of \app{} is therefore not to provide a permanently fixed dataset, but to support reproducible regeneration of labeled traces under new system configurations. This opens an important direction toward studying anomaly drift, detector transferability, and continual adaptation as MAS evolve.

\textbf{Enabling MAS Anomaly Detection Benchmarking.}
The primary goal of \app{} is to \emph{enable} systematic MAS anomaly detection, rather than to exhaustively cover the MAS design space. Today, MAS architectures vary widely in their workflows, agent organizations, communication protocols, tools, and memory mechanisms, while standardized practices for constructing and evaluating MAS AD benchmarks remain largely absent. We therefore first establish a set of benchmark-design principles: reproducible trace generation, configurable task construction, and deterministic fine-grained labeling, that make repeated AD development and evaluation possible. In this sense, \app{} serves as a foundation for defining how MAS AD benchmarks can be constructed and maintained as models and anomaly distributions evolve.

\textbf{Expanding the MAS Design Space.}
Broader MAS configurations represent different instantiations of the same benchmark-design principles. Changes in harness, workflow, agent organization, communication, tools, or memory may expose new anomaly surfaces and propagation patterns, but the principles of configurable task construction, oracle-verifiable behavior, refreshable traces, and deterministic labeling remain unchanged. Extending \app{} therefore amounts to defining a new MAS instantiation under the same paradigm, with its corresponding execution structure, anomaly granularity, and labeling oracles. An important future direction is to make \app{} more modular and automated, enabling new MAS components and design choices to be incorporated with minimal effort while automatically generating the corresponding workflows, oracles, and labels. 

\color{black}            
\section{Labeling Cost Analysis}
\label{app:label_cost}

We compare the annotation burden of existing trace analysis benchmarks in Table \ref{tab:benchmark_comparison} on a common per-trace basis,
using a representative hourly rate of $H = \$20$/hr where applicable.
Table~\ref{tab:cost} summarizes the estimated cost for each benchmark.

\begin{table*}[ht]
\centering
\caption{Estimated per-trace labeling cost across representative trace analysis benchmarks.
$H=\$20$/hr is used for human annotation estimates.
}
\label{tab:cost}
\small
\setlength{\tabcolsep}{8pt}
\begin{tabular}{lcr}
\toprule
\textbf{Benchmark} &
\textbf{Annotation method} &
\textbf{Est.\ cost/trace} \\
\midrule
TRAIL~\citep{Trail}
  & Human, 5-person pipeline
  & $\approx\$38$ \\

MAST~\citep{MAST}
  & LLM-as-judge (\texttt{o1}) + human QA
  & ${>}\$1$ \\

Who\&When~\citep{whoandWhen}
  & 3 experts, 3-round consensus
  & $\approx\$9.2$ \\

Silent Failures$^{*}$~\citep{SilentFailures}
  & Script + human GT per prompt$^{\dagger}$
  & small (unquantified) \\

AgenTracer~\citep{AgenTracer}
  & Counterfactual replay$^{\ddagger}$
  & unreported \\
\midrule
\textbf{\app{} (ours)}
  & Deterministic, generated ground-truth
  & $\mathbf{\approx\$0}$ \\
\bottomrule
\end{tabular}

\raggedright\small$\dagger$~Partially automated: cycles/errors are detected by script, but drift still requires human inspection.
$\ddagger$~Fully automated via counterfactual replay; only LLM API cost is incurred.
\end{table*}

\textbf{TRAIL~\citep{Trail}.}
Annotation follows a five-person human pipeline: one primary annotator spends an average of
35 minutes per trace (30\,min for GAIA, 40\,min for SWE-Bench), followed by four independent
verification rounds of 20 minutes each---totalling approximately 115 person-minutes per
trace~\citep{Trail}. At $H = \$20$/hr, this yields an estimated cost of \textbf{\$38 per trace}.

\textbf{MAST~\citep{MAST}.}
Failure modes are labeled by an LLM-as-a-Judge pipeline using OpenAI \texttt{o1}, at a
reported average API cost of \$1.80 per trace (ranging from \$0.37 to \$4.14 depending on the
MAS framework)~\citep{MAST}. A subset of traces additionally receives human annotation (HE and HA)
for quality assurance, though the corresponding human effort is not reported separately.
The total per-trace cost therefore \textbf{exceeds \$1}, with the human overhead unquantified.

\textbf{Who\&When~\citep{whoandWhen}.}
Three expert annotators independently label every trace across three consensus rounds
(independent labeling $\to$ uncertainty discussion $\to$ cross-validation). The total reported
annotation effort is 84.3 person-hours for 184 traces~\citep{whoandWhen}, corresponding to
approximately 27.5 minutes per trace per annotator. At $H = \$20$/hr, this yields an estimated
cost of \textbf{\$9.2 per trace}.

\textbf{Silent Failures~\citep{SilentFailures}.}
Cycle and error labels are assigned automatically by a rule-based script at negligible cost.
Drift labels require a domain expert to define the ground-truth expected trajectory once per
unique input prompt, after which labels are automatically propagated to all traces sharing that
prompt~\citep{SilentFailures}. Across 637 distinct prompts and 5{,}169 traces ($r \approx 8.1$
traces per prompt), this amortization substantially reduces the per-trace human burden,
though the exact annotation time per prompt is not reported. The per-trace cost is
\textbf{small but not precisely quantifiable}.

\textbf{AgenTracer~\citep{AgenTracer}.}
Labels are generated fully automatically via counterfactual replay and programmatic fault
injection, requiring no human annotator~\citep{AgenTracer}. An analyzer LLM (DeepSeek-R1) is
invoked once per candidate step until the decisive error is identified, so the API cost scales
with the depth of the error in the trace. No aggregate cost figure is reported by the authors;
the per-trace cost depends on trace length and the LLM backbone used, but is
\textbf{bounded by LLM API usage}.

\textbf{\app{} (ours).}
Ground-truth answers and labels are generated together with the puzzles and actions, requiring no human
annotator or LLM judge. The marginal cost of labeling a new trace is therefore
resulting in \textbf{effectively \$0} marginal cost per trace after environment construction.         
\section{Escape Room Design Details}
\label{app:room_setup}

\subsection{An Example of Escape Rooms.}
\label{app:room_example}

We show an example of an easy escape room with one puzzle from the
\textsc{LiveCodeBench} domain. The room contains three scenery items
(Items 1--3: a concrete wall, an oak table, and a Persian rug) that serve
as distractors with no task relevance. Item~4 is the \emph{real clue}: it
contains a chain of two syntactically valid Python programs and a hint
identifying a thermometer measured on the absolute temperature scale, i.e.,
Kelvin. Item~7 is a \emph{fake clue}: its first Python program is broken
because the expression \texttt{offers[i][2)} has mismatched brackets and
therefore would not run. Item~5 is the \emph{real instrument}, a glass
thermometer reading \texttt{88 celsius}. Item~6 is a \emph{fake instrument}:
although it is described as being in a comfortable room, it reads
\texttt{-218 celsius}, which is inconsistent with the room state and should
be discarded.

\begin{tcolorbox}[title=\textbf{An escape room (easy) with one LiveCodeBench puzzle}, breakable]
\footnotesize
\begin{Verbatim}[breaklines=true, breakanywhere=true]
You now see the following objects in the room:
  [Item 1] A plain concrete wall. There are a few hairline cracks running across it, but nothing you can pry open.
  [Item 2] A sturdy oak table in the center of the room. Its surface is scratched with faint marks that might once have meant something.
  [Item 3] A faded Persian rug covering most of the floor. Lifting the corner reveals nothing but dust underneath.
  [Item 4] A note on the floor which may contains a clue. It reads "This clue is a chain of two Python programs; run them in order.

FIRST program - run it on its call and let a = the value it returns:

def maxSubarrays(nums: List[int]) -> int:
    n = len(nums)
    mn = nums[0]
    for num in nums:
        mn &= num
    if mn:
        return 1
    res, cur = 0, nums[0]
    for i in range(1, n):
        if cur == mn:
            res += 1
            cur = nums[i]
        cur &= nums[i]
    if cur == mn:
        res += 1
    return res

Call: maxSubarrays(nums = [1, 0, 2, 0, 1, 2])

SECOND program - replace the argument `z` in its call with a (the value from the FIRST program), then run it. The value it returns is the clue's result:

def longestString(x: int, y: int, z: int) -> int:
    if x > y:
        return 2 * (y + min(y + 1, x) + z)
    else:
        return 2 * (x + min(x + 1, y) + z)

Call: longestString(x=3, y=2, z=a)"
Hint: Something with a thin glass tube and silver liquid awaits the answer, counted in the absolute scale that starts at nothing.
  [Item 5] A glass thermometer that measures temperature. It reads 88 celsius.
  [Item 6] A glass thermometer mounted on the wall. The room feels perfectly comfortable. It reads -218 celsius.
  [Item 7] A note on the floor which may contains a clue. It reads "This clue is a chain of two Python programs; run them in order.

FIRST program - run it on its call and let a = the value it returns:

def maximizeTheProfit(n: int, offers: List[List[int]]) -> int:
    dp = [0] * (n + 1)
    offers.sort()
    i = 0
    for r in range(n + 1):
        dp[r] = max(dp[r], dp[r - 1])
        while i < len(offers) and offers[i][0] <= r:
            dp[offers[i][1] + 1] = max(dp[offers[i][1] + 1], dp[offers[i][0]] + offers[i][2)
            i += 1
    return dp[-1]

Call: maximizeTheProfit(n = 5, offers = [[0, 0, 1], [0, 2, 10], [1, 3, 2]])

SECOND program - replace the argument `k` in its call with a (the value from the FIRST program), then run it. The value it returns is the clue's result:

def maximumBeauty(nums: List[int], k: int) -> int:
    nums.sort()
    j = 0
    ans = 0
    for i in range(len(nums)):
        while j < len(nums) and nums[j] - nums[i] <= 2 * k:
            j += 1
        ans = max(ans, j - i)
    return ans

Call: maximumBeauty(nums=[4, 6, 1, 2], k=a)
Hint: An instrument of orientation depends on what you find, counted in the unit where one means a full circle."
\end{Verbatim}
\end{tcolorbox}

To solve this puzzle, one must first identify Item~4 as the real clue by
checking code validity: both programs in Item~4 are runnable, whereas
Item~7 contains a syntax error in its first program. The first program in
Item~4 returns $a=3$ on the given call. Substituting this into the second
program gives \texttt{longestString(x=3, y=2, z=3)}, which returns $16$.
Following the benchmark instruction, the raw code result is transformed by
adding $1{,}000{,}000$, yielding a final delta of $1{,}000{,}016$ Kelvin.
Item~5 is then identified as the correct instrument. Since a temperature
delta of $1{,}000{,}016$ Kelvin is equal to a delta of $1{,}000{,}016$
Celsius, applying it to the thermometer reading of \texttt{88 celsius}
yields $88 + 1{,}000{,}016 = 1{,}000{,}104$ Celsius. The puzzle is solved
with the correct adjusted reading, and the room state reflects this by
removing the clue and the instrument. A more detailed action and labeling
description is provided in Appendix~\ref{app:label}.
\subsection{Puzzle Generation}  
  \textbf{Instruments.} Each escape room instance is procedurally generated as a set of interconnected puzzles, 
  where each puzzle is anchored by one of four physical instrument types: \textit{thermometer}             
  ($[-20, 120]$\textdegree, convertible to units of either Celsius, Fahrenheit, or Kelvin;  
  \textit{compass} ($[0, 359]$\textdegree; convertible to units including degree, radians or turns), \textit{clock} ($[0, 86399]$\,s; convertible to seconds, minutes or hours, with 24-hour wrapping), and \textit{scale} ($[1, 100]$\,kg; convertible to grams, pounds, or ounces). 
  
   \textbf{Hint.} For each puzzle, we sample a math/coding word problem from the GSM-Hard dataset~\citep{GSM-hard} or LiveCodeBench-execution~\citep{jain2025livecodebench} as the \emph{clue}, whose numerical answer serves as a delta to be applied to the  instrument's current reading after appropriate unit conversion. The ground-truth answer is thus $s' = \text{wrap}(s + \Delta_{\text{converted}})$, where $s$ is the instrument's current state and $\Delta_{\text{converted}}$ is the clue answer converted into the instrument's native unit. A poetic hint (e.g., ``\textit{the scale where water freezes at zero}'') links each clue to its target instrument without revealing the answer directly.         
  
  \textbf{Traps and Distractions.} To increase difficulty, we introduce three layers of \emph{traps}:                     
  (1)~\textbf{fake clues} ($n_{\text{fake}} \in [1, 5]$), which contain misleading hints referencing non-existent instruments;                        
  (2)~\textbf{fake items} ($n_{\text{fake}} \in [1, 5]$), which share the same instrument type as the real item but exhibit telltale signs such as out-of-range readings or descriptions of broken/non-functional states (e.g., ``\textit{the mercury level hasn't shifted in hours}''); and (3)~\textbf{item-value description traps}, which embed adversarial text in the real item's description, suggesting that the current reading is already the correct answer.        Additionally, $n_{\text{scenery}}$ irrelevant decorative objects are placed in the room as distractors.

\subsection{Extreme Room Design}
\label{app:room_setup:nightmare}

\textbf{Overall.} Extreme rooms extend the standard 
EscapeRoom design with explicit inter-puzzle dependencies, 
requiring agents to determine a valid solving order. Each room 
contains four puzzles, one per instrument; solving a puzzle 
unlocks some instruments and locks others, so the agent must 
plan the order such that all four puzzles eventually become 
solvable. We formalise this as a STRIPS planning 
instance~\citep{fikes1971strips}, a classical framework also 
underlying the Blocks World domain~\citep{winograd1971procedures}.

\paragraph{A Concrete Example.}
A Extreme room with four puzzles ($P_{\textsc{clock}}$, 
$P_{\textsc{thermometer}}$, $P_{\textsc{compass}}$, 
$P_{\textsc{scale}}$) starts with three instruments unlocked 
(Thermometer, Compass, Scale) and Clock locked. Solving each 
puzzle modifies the lock states of others; only \emph{two} 
valid orderings exist:
\begin{itemize}
    \item Path A: $P_{\textsc{compass}} \to P_{\textsc{scale}} \to P_{\textsc{thermometer}} \to P_{\textsc{clock}}$
    \item Path B: $P_{\textsc{compass}} \to P_{\textsc{clock}} \to P_{\textsc{scale}} \to P_{\textsc{thermometer}}$
\end{itemize}
Both require starting with $P_{\textsc{compass}}$. Four other 
orderings lead to \textit{deadends}, with the deepest making three 
seemingly valid moves before the room becomes unsolvable. The 
planner is given full state information at every step, with 
nothing hidden. A globally reasoning planner can trace the 
consequences of each choice and identify $P_{\textsc{compass}}$ 
as the only safe first move; a greedy, one-step-lookahead 
planner naturally falls into deadends---for example, choosing 
$P_{\textsc{scale}}$ at step 0 looks harmless (unlocks 
Thermometer, locks nothing) but creates a downstream constraint 
that eventually makes the room unsolvable. The deadend emerges 
\emph{organically} from short-horizon planning against a DAG 
requiring global foresight.

\textbf{DAG Sampling. }The lock/unlock effects induce a directed dependency graph over puzzles. We enumerate all $4! = 24$ possible orderings via depth-first search over the state space. A \emph{deadend} occurs when all remaining instruments are locked before the goal is reached. A \emph{trap first move} is a first-step choice after which no solution path exists regardless of subsequent decisions.

Extreme DAGs are generated by rejection sampling and can be configured with different difficulty levels. The current configuration is: (1) at least one complete solution path exists; (2) at least two deadend paths exist; (3) at least one deadend path has depth $\geq 2$, ensuring traps are not trivially detectable at the first step; and (4) at least two first-step choices are guaranteed traps, with at least one safe choice remaining --- bounding the random success rate at the first step to at most $1/3$. These constraints are verified exactly by exhaustive graph enumeration, which is tractable given the fixed state space of $|\mathcal{L}| \leq 4$ instruments and four puzzles. The full configuration space of $4 \times 15^4 = 202{,}500$ possible DAGs yields \textbf{2,376} structurally valid instances satisfying all constraints (Appendix~\ref{app:instance-space}).

\subsection{Task Instance Space}
\label{app:instance-space}
We calculate the task instance space with GSM-hard math questions as an example. Each puzzle instance is determined by the independent combination of the following
factors: the math questions (1,319 choices);
the instrument type (one of four: clock, compass, thermometer, or scale);
the initial instrument reading ($|\mathcal{S}_i|$, ranging from 100 to 86,400
states depending on the instrument); and the delta unit ($|\mathcal{U}_i|$,
3--4 choices per instrument type).
For Extreme rooms, an additional dependency layer contributes 2,376 structurally
distinct DAG configurations (Appendix~\ref{app:room_setup:nightmare}). These factors only cover the components required to unlock each puzzle/room; we conservatively exclude additional variation from distractors and traps.

Since each puzzle in a room is configured independently, the total instance space is the product of per-puzzle factors across all $n_d$ puzzles:
$\mathcal{I}(d)=\left(|\mathcal{P}|\cdot|\mathcal{S}_i|\cdot|\mathcal{U}_i|\right)^{n_d}$,
where $|\mathcal{P}|=1{,}319$.
Because the instrument type varies per puzzle, the instance space is bounded by
the least flexible instrument (Scale: $100 \times 4 = 400$ configurations) and
the most flexible (Clock: $86{,}400 \times 3 = 259{,}200$ configurations),
giving a per-puzzle base factor between $1{,}319 \times 400 \approx 5\times10^5$
and $1{,}319 \times 259{,}200 \approx 3.4\times10^8$.
The resulting instance space ranges are:

\textbf{Easy} ($n=1$):
        $[5\times10^5,\ 3.4\times10^8] \approx [10^{6},\ 10^{9}]$
        
\textbf{Medium} ($n=2$):
        $[(5\times10^5)^2,\ (3.4\times10^8)^2] \approx [10^{11},\ 10^{17}]$
        
\textbf{Hard} ($n=3$):
        $[(5\times10^5)^3,\ (3.4\times10^8)^3] \approx [10^{17},\ 10^{26}]$
        
\textbf{Extreme} ($n=4$, $\times\,2{,}376$ DAG variants):
        $[(5\times10^5)^4 \times 2376,\ (3.4\times10^8)^4 \times 2376]
         \approx [10^{26},\ 10^{37}]$
         
Table~\ref{tab:instance-space} summarises these ranges across all difficulty levels.

\begin{table}[t]
\centering
\caption{Task instance space per difficulty level. Each range reflects the least
flexible instrument (Scale, $100\times4$ configurations) as the lower bound and
the most flexible (Clock, $86{,}400\times3$) as the upper bound.
Extreme additionally multiplies by 2,376 valid DAG configurations.}
\small
\begin{tabular}{lcccr}
\toprule
\textbf{Difficulty} & \textbf{Puzzles} & \textbf{Fake Items/Clues}
  & \textbf{DAG Variants} & \textbf{Instance Space} \\
\midrule
Easy      & 1 & 1 & ---   & $[10^{6},\ 10^{9}]$  \\
Medium    & 2 & 3 & ---   & $[10^{11},\ 10^{17}]$ \\
Hard      & 3 & 5 & ---   & $[10^{17},\ 10^{26}]$ \\
Extreme & 4 & 3 & 2,376 & $[10^{26},\ 10^{37}]$ \\
\bottomrule
\end{tabular}
\label{tab:instance-space}
\end{table}

The combinatorial scale across all difficulty levels renders memorisation-based
solutions intractable and ensures that each of the evaluation traces
constitutes a statistically independent problem instance.

\subsection{Action Labeling}
\label{app:label}

Each action is labeled as \emph{success} or \emph{failure} by comparing the agent's structured output with the pre-generated ground truth. For failed actions, an error-code  is further assigned during labeling. The action-specific labeling criteria are as follows.

\textbf{\textsc{Select\_Puzzle} (Extreme only).}
The agent must select a puzzle that is currently available (i.e., its instrument is unlocked and the puzzle has not yet been solved). The action is labeled a failure if the selected puzzle has already been solved (error-code: \texttt{already\_solved}), if its instrument is currently locked (\texttt{puzzle\_locked}), if the puzzle ID does not exist in the room (\texttt{puzzle\_not\_found}), or if the selection leads to an irrecoverable deadend state (\texttt{leads\_to\_deadend}).

\textbf{\textsc{Observe\_Clue}.}
The agent must select the item ID corresponding to the real clue. The action is labeled a failure if the selected item ID does not match the ground-truth clue ID (\texttt{clue\_select\_wrong}).

\textbf{\textsc{Solve\_Clue}.}
The agent must produce both the correct numeric answer and the correct delta unit. The action is labeled a failure if the numeric value deviates from the ground-truth answer beyond a relative tolerance of $10^{-4}$ (\texttt{math\_value\_wrong}), or if the reported unit does not match the ground-truth unit (\texttt{math\_unit\_wrong}). Both checks are applied independently.

\textbf{\textsc{Observe\_Instrument}.}
Given an instrument type from \textsc{Solve\_Clue}, the agent must identify the correct instrument from the distractors. It also needs to report its current reading. The action is labeled a failure if the reported instrument ID does not match the ground-truth ID (\texttt{instrument\_wrong}), or if the reported state deviates from the ground-truth reading beyond a relative tolerance of $10^{-4}$ (\texttt{instrument\_state\_wrong}).

\textbf{\textsc{Apply\_Delta}.}
The agent must compute the correct final reading after applying the delta under the correct unit conversion. The action is labeled a failure if the final answer does not match the ground-truth result (\texttt{apply\_delta\_wrong}). 

\textbf{\textsc{Tool\_Call} (Tool-enabled setting only).}
When unit-conversion tools are enabled and a conversion is required, the agent
must issue a valid structured tool call before producing the final
\textsc{Apply\_Delta} answer. The action is labeled a failure if the required
conversion is not called (\texttt{TC\_REQUIRED\_NOT\_CALLED}), if the selected
tool is incompatible with the required conversion (\texttt{TC\_WRONG\_TOOL}), if
the tool arguments use an incorrect value or source/target unit
(\texttt{TC\_WRONG\_ARGUMENTS}), if the requested conversion crosses
incompatible unit dimensions (\texttt{TC\_DIMENSION\_MISMATCH}), if the tool call
is malformed or otherwise invalid (\texttt{TC\_INVALID\_REQUEST}), or if the
agent omits the required auditable tool-call message
(\texttt{TC\_MISSING\_MESSAGE}). These tool-call failures are counted as
action-level anomalies; in the failure-mode analysis, their downstream effect is
captured through the parent \textsc{Apply\_Delta} action.

\textbf{\textsc{Observe\_Puzzle}.}
The agent must correctly assess whether the puzzle has been solved based on the current room state. The action is labeled a failure if the agent's judgment disagrees with the actual puzzle state (\texttt{observer\_disagrees}).    
\section{MAS Design Details}
\label{app:mas}


\subsection{MAS Harness}
\label{app:harness}

Our harness implements a sequential, role-based multi-agent workflow comprising an observer, a clue solver, an item manager, and a planner for rooms with puzzle dependencies. Each role can perform multiple related actions. The orchestrator constructs action-specific prompts from the current environment state and upstream messages. Agents return JSON objects containing a natural-language \texttt{message} and \texttt{structured} outputs; these outputs are passed to downstream agents together with the message as a \texttt{reasoning} field. Submitted instrument readings trigger environment updates, which inform subsequent observations and planning decisions. The harness records action inputs and outputs, agent identities, LLM-call token usage and latency, and tool-call metadata, and annotates actions against the room's oracle information. This scaffold provides a controlled setting for evaluating action-level reasoning and error propagation.

\paragraph{Tool calls.}
The harness supports both Tool and No-Tool settings under the same role structure and action pipeline. In the No-Tool setting, the agent performs unit conversion itself within \texttt{APPLY\_DELTA}. In the Tool setting, it can request unit conversion tools through \texttt{convert\_time}, \texttt{convert\_temperature\_delta}, \texttt{convert\_angle}, and \texttt{convert\_mass}. 
A request specifies the tool name and the arguments \texttt{value}, \texttt{from\_unit}, and \texttt{to\_unit}. The harness returns the result or error to the agent, which can issue another request, up to three calls, or submit its final answer. Oracle-based checks identify missing required calls, incorrect tool choices or arguments, and incorrect final readings.

\color{black}

\begingroup
\tcbset{breakable}

\subsection{Action and Agent Prompts}
\label{app:prompts}

We provide representative prompts here for reference; the complete prompt
templates are available in our code repository.

\begin{tcolorbox}[title=\textbf{Agent Prompts}]
\footnotesize
\begin{verbatim}
Observer:
  You are the Observer in an escape room. You examine objects in the
  room and distinguish real instruments and clues from fakes.
  Fake items have tells: broken/damaged descriptions, or readings that
  are physically impossible for what they claim to be.
  Fake clues reference nonsensical instruments or units that don't
  exist. Real clues contain a solvable math problem and a hint
  pointing to a specific type of instrument and unit.
  Always respond with a single valid JSON object — no markdown,
  no extra text.
Clue Solver:
  You are the Clue Solver. You receive a math word problem extracted
  from a clue found in the room. Solve it step by step and return
  the numerical answer.
  Always respond with a single valid JSON object — no markdown,
  no extra text.
Item Manager:
  You are the Item Manager. You receive an instrument's current
  reading, a delta value, and a unit. Compute the new reading after
  applying the delta. Pay attention to unit conversions.
  Always respond with a single valid JSON object — no markdown,
  no extra text.
Planner:
  You are the Planner in an escape room with multiple puzzles.
  You must choose which puzzle to solve next based on item lock
  states and DAG effects. A wrong ordering may permanently lock
  required items.
  Always respond with a single valid JSON object — no markdown,
  no extra text.
\end{verbatim}
\end{tcolorbox}

\begin{tcolorbox}[
  title=\textbf{Action: APPLY\_DELTA}
        \textnormal{(agent: item\_manager)}
]
\footnotesize
\begin{verbatim}
You are operating an instrument in an escape room.

[Previous input — OBSERVE_INSTRU structured JSON]:
  "item_type" — instrument type
  "item_state" — current reading of the instrument (number)
  "item_unit" — unit of the current reading
  "delta" — the value to add to the reading (from the clue solver)
  "delta_unit" — the unit of the delta
IMPORTANT: use ONLY the values given in the JSON. Do not substitute
or guess.

Your task: compute the new instrument reading after applying the delta.

Step-by-step instructions:
  1. Extract the instrument's current reading, unit, delta value,
     and delta unit from the input above
  2. If the delta unit differs from the instrument's unit, convert
     the delta first
  3. Add the converted delta to the current reading
  4. Apply wrapping if needed (clocks wrap at 24h, compasses
     at 360 degrees)

Return a JSON object with exactly these two keys:

"message": "<step-by-step calculation showing unit conversion and
            final result>",
"structured": {
  "answer": <final reading. Clocks: "HH:MM:SS"; others: integer>,
  "converted_delta": <delta after unit conversion, in instrument's
                      unit>,
  "converted_unit": "<unit after conversion — same as instrument
                      unit>"
}
\end{verbatim}
\end{tcolorbox}

\begin{tcolorbox}[
  title=\textbf{Action: OBSERVE\_CLUE}
        \textnormal{(agent: observer)}
]
\footnotesize
\begin{verbatim}
You are observing an escape room. Here is everything you can see:

{room_desc}

Your task: identify the ONE real clue among the notes on the floor.

A REAL clue has both:
  1. A math problem (a question involving numbers and calculation)
  2. A hint that describes a real instrument and a unit of measurement

Select the single real clue.

Return a JSON object with exactly these two keys:

"message": "<Include the FULL TEXT of the math problem and the FULL
            TEXT of the hint. The next agent needs the exact problem
            to solve it.>",
"structured": {
  "item_id": <integer, the item id of the clue>,
  "clue":    "<exact full text of the clue>"
}
\end{verbatim}
\end{tcolorbox}

\begin{tcolorbox}[
  title=\textbf{Action: OBSERVE\_INSTRU}
        \textnormal{(agent: observer)}]
\footnotesize
\begin{verbatim}
You are observing an escape room. Here is everything you can see:

{room_desc}

[Previous input — SOLVE_CLUE structured JSON]:
  "answer" — the numerical delta value computed from the clue
  "unit"   — the unit of that delta

Your task: identify the ONE real instrument that the clue's hint
refers to.
A REAL instrument:
  - Has a physically valid reading and a working description
  - Matches the instrument type described in the clue
Select the single real instrument.

Return a JSON object with exactly these two keys:
"message": "<Include: (1) the instrument you found and its current
            reading with unit, (2) the delta value and delta unit
            from the solver's input above.>",
"structured": {
  "item_index": <integer, 1-based index of the instrument>,
  "item_type":  "<one of: {valid_types}>",
  "item_state": <current reading as a number>,
  "item_unit":  "<unit of the current reading>",
  "delta":      <number — the delta value from the solver>,
  "delta_unit": "<unit of the delta from the solver>"
}
\end{verbatim}
\end{tcolorbox}

\begin{tcolorbox}[
  title=\textbf{Action: OBSERVE\_PUZZLE}
        \textnormal{(agent: observer)}
]
\footnotesize
\begin{verbatim}
You are observing an escape room. Here is everything you can see:

{room_desc}

[Previous input — OBSERVE_CLUE and OBSERVE_INSTUR structured JSONs]

Your task: determine whether the current puzzle has been solved by
comparing the current room contents against your previous observation.

Step 1 — Inventory changes:
  List every item that DISAPPEARED and every item that is NEW.
Step 2 — Check the clue and instrument you previously identified:
  - Is the clue you identified still present in the room?
  - Is the instrument you identified still present in the room?
Step 3 — Determine solved:
  The puzzle is solved ONLY if BOTH the previously identified clue
  AND instrument have DISAPPEARED. If either is still present,
  the puzzle is NOT solved.

Return a JSON object with exactly these two keys:

"message": "<Step 1: list disappeared/new items. Step 2: check
            presence. Step 3: conclusion.>",
"structured": {
  "solved":       <true ONLY if both clue and item are gone>,
  "clue_present": <true if the previously identified clue is visible>,
  "item_present": <true if the previously identified instrument is
                   visible>
}
\end{verbatim}
\end{tcolorbox}

\begin{tcolorbox}[
  title=\textbf{Action: SELECT\_PUZZLE}
        \textnormal{(agent: planner)}
]
\footnotesize
\begin{verbatim}
You are a planner in an escape room with multiple puzzles.
Each puzzle is tied to an item. Items can be UNLOCKED or LOCKED.
You can only solve puzzles whose item is UNLOCKED.
Solving a puzzle may unlock or lock other items — choose wisely.

{planning_desc}

Your task: choose ONE puzzle to solve next.
Consider the unlock/lock effects carefully — a wrong ordering
may permanently lock required items (deadend).

Return a JSON object with exactly these two keys:

"message": "<your reasoning about which puzzle to solve and why>",
"structured": {
  "puzzle_id": "<the puzzle_id you choose to solve next>",
  "item_type": "<item type of the chosen puzzle —
                one of: {valid_types}>"
}
\end{verbatim}
\end{tcolorbox}

\begin{tcolorbox}[
  title=\textbf{Action: SOLVE\_CLUE}
        \textnormal{(agent: clue\_solver)}
]
\footnotesize
\begin{verbatim}
You are a clue solver in an escape room.

[Previous input — OBSERVE_CLUE structured JSON]:
  "item_id" — the clue's item id
  "clue" — the full text of the clue (math problem + instrument hint)

Your task: extract the math problem from the input above and solve it.
Your numerical answer will later be used as a delta with a specified
unit to adjust an instrument in the room.

Instructions:
  - Solve step by step
  - Your final answer must be a single number
  - The input above tells you which unit your answer should be
    expressed in — include that unit in your structured output
  - Do NOT apply the delta to any instrument yet

Return a JSON object with exactly these two keys:

"message": "<step-by-step solution of the math problem,
             show all work>",
"structured": {
  "answer": <number — the final solution>,
  "unit":   "<unit string — must be one of: {valid_units}>"
}
\end{verbatim}
\end{tcolorbox}

\endgroup

\subsection{Computational Cost of \appfull{}}
\label{app:trace_cost}

\textbf{Cost of Generating \appfull{}.}
Table~\ref{tab:api-cost} reports the API cost of generating
\appfull{} across five SOTA LLM backbones, aggregated over our
full sampling sweep (two clue domains -- GSM-Hard and
LiveCodeBench -- three sampling temperatures $\{0.0, 0.3, 0.6\}$,
and with/without the unit-conversion tool). The total cost is
\$394.32 for 
\fullsize{}
traces (averaging \$0.076 per trace). Cost varies
primarily with output token volume and price per million tokens:
Claude Sonnet 4.6 (\$146.63) is the most expensive backbone due to
its \$15/1M output rate combined with the highest aggregate token
volume, while DeepSeek-R1 (\$39.57) is the cheapest despite
producing the most output tokens (13.05M), owing to its
\$0.55/\$2.19 per-1M rate. Claude Opus 4.8 (\$93.44) currently
covers only the temperature-0.0 configurations (400 traces) and
is thus not directly comparable on a per-backbone total, though at
\$0.234/trace it is the most expensive backbone per trace.
\begin{table}[h]
\centering
\caption{API cost for generating \appfull{} across five SOTA LLM
backbones, aggregated over all sampled configurations (domains,
temperatures, and tool/no-tool variants). Prices are based on
official API rates at experiment time. Claude Opus 4.8 currently
covers only the default temperature configuration (400 traces); all
other backbones cover 1{,}200 traces each.}
\label{tab:api-cost}
\resizebox{\textwidth}{!}{%
\begin{tabular}{l c c c c c c}
\toprule
\textbf{Model} & \textbf{Traces} & \textbf{API Calls} & \textbf{Input Tokens} & \textbf{Output Tokens} & \textbf{Price (in/out per 1M)} & \textbf{Total Cost} \\
\midrule
DeepSeek-R1             & 1{,}200 & 10{,}267 & 19.96M & 13.05M & \$0.55 / \$2.19  & \$39.57  \\
Claude Sonnet 4.6       & 1{,}200 & 10{,}329 & 20.26M & 5.72M  & \$3.00 / \$15.00 & \$146.63 \\
Claude Opus 4.8         & 400     & 4{,}025  & 10.27M & 1.68M  & \$5.00 / \$25.00 & \$93.44  \\
GPT-4.1                 & 1{,}200 & 7{,}605  & 13.27M & 2.81M  & \$2.00 / \$8.00  & \$48.98  \\
GPT-5.4                 & 1{,}200 & 8{,}445  & 14.01M & 2.04M  & \$2.50 / \$15.00 & \$65.71  \\

\midrule
\textbf{Total} & \textbf{5{,}200} & \textbf{40{,}671} & \textbf{77.77M} & \textbf{25.31M} & -- & \textbf{\$394.32} \\
\bottomrule
\end{tabular}%
}
\end{table}            
\section{Additional Anomaly Profiling}
\label{app:more_profile}




\textbf{Mapping \app{} Anomalies to Failure Modes.}
To support interpretable anomaly analysis, \app{} automatically maps each failed action  (by their error code, Appendix \ref{app:label}) to a MAST~\citep{MAST} failure mode, through rule-based classification. Table \ref{tab:fm-mapping} summarizes the rules. 

\textbf{Extending the MAST Taxonomy.}
MAST's original taxonomy (FC1--FC3) mainly covers system-level failures, including prompt design, agent coordination, and output verification. However, \app{} reveals two additional failure families that are central to MAS AD. We therefore extend MAST with \textbf{FC4~(Agent Capability)} and \textbf{FC5~(Error Propagation)}. FC4 captures single-agent capability failures independent of orchestration, including \textbf{FM-4.1~Deception Susceptibility}, where an agent follows a distractor, and \textbf{FM-4.2~Reasoning Error}, where arithmetic or logical reasoning fails. \textbf{FM-5} captures  downstream action failures due to upstream failures. These categories are orthogonal to the coordination and prompt-design failures targeted by MAST's FC1--FC3.

\begin{figure}[h]
\vspace{-10pt}
    \centering
    \includegraphics[width=0.99\linewidth]{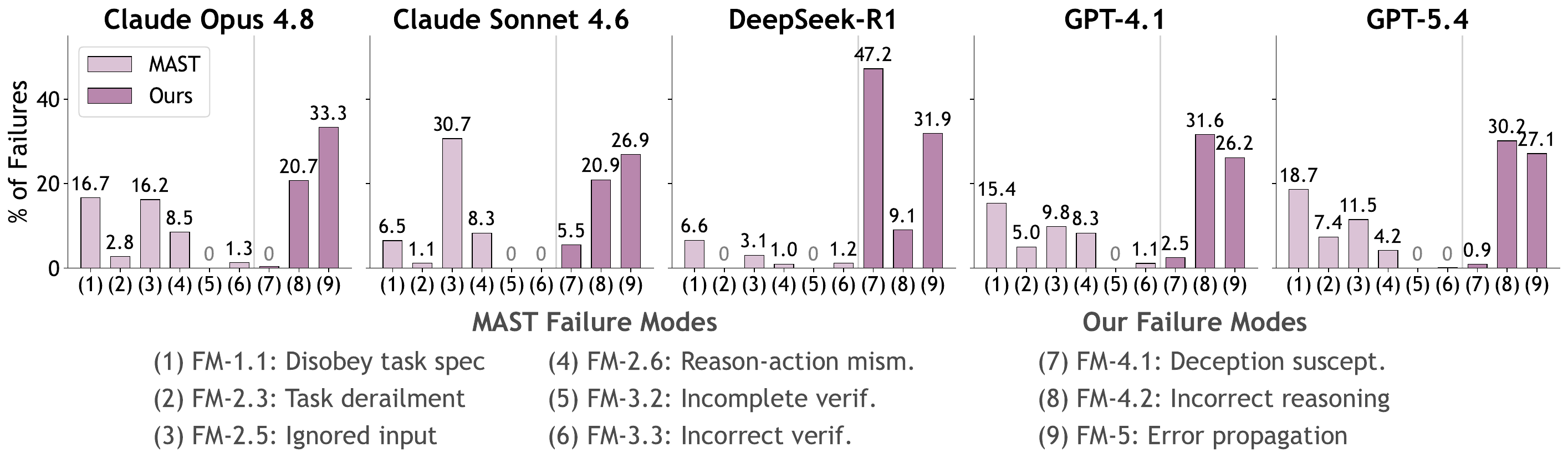}
    \caption{Distribution of failures across five LLM backbones, broken down by MAST failure modes (FM-1.1--FM-3.3, light bars) and \app{}'s extended failure modes (FM-4.1, FM-4.2 and FM-5, dark bars).  FM-4.2 (Reasoning Error) and FM-5 (Error Propagation) dominate failures across all backbones.}
    \label{fig:fm_by_model}
\end{figure}

Figure~\ref{fig:fm_by_model} shows the resulting failure-mode distribution across LLM backbones. FM-4.2 and FM-5 dominate failures across all backbones, indicating that reasoning failures and error propagation are core failure patterns in \app{}.

\begin{table}[t]
\centering
\caption{
Deterministic mapping from \app{} action-level error codes to MAST failure modes. 
Cascade rule: if any upstream action on the same puzzle already failed, the downstream failure is labeled as \textbf{FM-5~Error Propagation}.
}
\label{tab:fm-mapping}
\resizebox{\linewidth}{!}{%
\renewcommand{\arraystretch}{1.18}
\begin{tabular}{l l l}
\toprule
\textbf{Action} & \textbf{Error code} & \textbf{Failure mode} \\
\midrule
\multirow{4}{*}{\textsc{Select\_Puzzle}}
& \texttt{puzzle\_locked} & FM-1.1~~Disobey task spec. \\
& \texttt{already\_solved} & FM-1.3~~Step repetition \\
& \texttt{leads\_to\_deadend} & FM-2.3~~Task derailment \\
& \texttt{puzzle\_not\_found} & FM-2.6~~Reason--action mismatch \\
\midrule
\multirow{2}{*}{\textsc{Observe\_Clue}}
& \texttt{clue\_select\_wrong}~(GT id $\in$ output reasoning content) & FM-2.6~~Reason--action mismatch \\
& \texttt{clue\_select\_wrong}~(GT id $\notin$ output reasoning content) & FM-4.1~~Deception suscept. \\
\midrule
\multirow{3}{*}{\textsc{Solve\_Clue}}
& \texttt{math\_unit\_wrong}~(value ok) & FM-1.1~~Disobey task spec. \\
& \texttt{math\_value\_wrong},~GT $\in$ output reasoning content & FM-2.6~~Reason--action mismatch \\
& \texttt{math\_value\_wrong},~GT $\notin$ output reasoning content & FM-4.2~~Reasoning error \\
\midrule
\multirow{3}{*}{\textsc{Observe\_Instrument}}
& \texttt{instrument\_state\_wrong}~(id ok) & FM-2.5~~Ignored input \\
& \texttt{instrument\_wrong},~GT $\in$ output reasoning content & FM-2.6~~Reason--action mismatch \\
& \texttt{instrument\_wrong},~GT $\notin$ output reasoning content & FM-4.1~~Deception suscept. \\
\midrule
\multirow{3}{*}{\textsc{Tool\_Call}}
& \texttt{TC\_REQUIRED\_NOT\_CALLED}, \texttt{TC\_DIMENSION\_MISMATCH}, \texttt{TC\_INVALID\_REQUEST} & FM-1.1~~Disobey task spec. \\
& \texttt{TC\_WRONG\_ARGUMENTS}, \texttt{TC\_WRONG\_TOOL} & FM-2.5~~Ignored input \\
& \texttt{TC\_MISSING\_MESSAGE}~only & FM-3.2~~Incomplete verification \\
\midrule
\multirow{2}{*}{\textsc{Apply\_Delta}}
& \texttt{apply\_delta\_wrong},~upstream values used & FM-4.2~~Reasoning error \\
& \texttt{apply\_delta\_wrong},~upstream values absent & FM-2.5~~Ignored input \\
\midrule
\textsc{Observe\_Puzzle}
& \texttt{observer\_disagrees} & FM-3.3~~Incorrect verif. \\
\midrule
\textit{any downstream}
& \textit{cascade from upstream failure} & \textbf{FM-5~~Error propagation} \\
\bottomrule
\end{tabular}%
}
\end{table}     
\newpage
\section{Experimental Setup Details}
\label{app:methods}
We organize the anomaly detection algorithms in \app{} in the following four supervision levels.

\subsection{Data Processing}
\label{app:data_processing}

For all methods, we randomly selected 80\% of the actions for training using stratified sampling, with the remaining 20\% reserved for testing. Each random seed produces a different combination of samples. The test set is consistent across all methods, while the training set differs depending on the level of supervision.

\noindent\textbf{Tabular Methods.} We convert each action trace into a tabular feature vector through the following steps. First, we clean raw LLM reasoning outputs by removing markdown artifacts, special characters, and normalizing whitespace. We then extract statistical features for each trace, including: (i) LLM call statistics (duration mean/min/max/std), and (ii) token consumption (input/output tokens mean/min/max/std). For semantic information, to ensure the previous context of an action, we concatenate all action outputs up to the current action and embed them using a pre-trained sentence transformer (all-MiniLM-L6-v2, 384-dim), as its contextual representation.

\noindent\textbf{Graph-based Methods.} For graph-based anomaly detection, we construct heterogeneous graphs from execution traces. We extract directed edges from the execution sequence: if Agent A is followed by Agent B, we add edge $A \rightarrow B$. Each agent node stores its embedded system prompt as a node attribute, while edges store embedded LLM reasoning outputs passed to the next agent, along with call duration metadata.

\noindent\textbf{MAS-specific Methods.} G-Safeguard and BlindGuard build directly on the graph representation described above.

\noindent\textbf{LLM-based Detectors.} We leverage a open-source LLM (Gemma-4) as an anomaly detector. The model receives complete execution traces in JSON format, with a detailed system prompt, motivated by MAST \citep{MAST}, that provides task context, describes the multi-agent system components. Anomaly definition prompts are revised based on MAST~\citep{MAST} LLM-as-a-judge prompts.

\newpage
\subsection{Method Details}
\label{app:method_detail}

\noindent\textbf{Unsupervised Methods.} Require no labels for both normal and abnormal data. 
\begin{enumerate}
    \item \textbf{Isolation Forest}~\citep{liu2008isolation} is an ensemble-based anomaly detection method that isolates observations by randomly selecting a feature and splitting the data. Anomalies are identified as instances that require fewer splits to be isolated, resulting in shorter average path lengths in the ensemble of isolation trees.
    
    \item\textbf{KNN}~\citep{ramaswamy2000efficient} detects anomalies by computing the distance of each data point to its $k$-nearest neighbors. Points with large distances to their neighbors are considered anomalous, as they lie in sparse regions of the feature space.

    \item\textbf{LOF}~\citep{breunig2000lof} (Local Outlier Factor) measures the local density deviation of a data point with respect to its neighbors. Anomalies are identified as points whose local density is significantly lower than that of their neighbors, indicating they reside in sparser regions.

    \item \textbf{Gemma4-E2B (ZS)}, \textbf{Gemma4-E4B (ZS)}, \textbf{Gemma4-31B (ZS)}, and \textbf{Gemma4-26B-A4B (ZS)}~\citep{gemma4_deepmind_2026} are open-source LLMs from the Gemma4 family, applied in a zero-shot setting. We choose Gemma for two reasons: (i) its open-source release supports reproducibility, and (ii) its multiple size variants (2B--31B parameters) enable us to systematically study how LLM-based detection scales with model capacity. They are prompted to assess whether a given instance deviates from normal behavior based purely on the model's pretrained knowledge. The prompts are provided here \url{https://drive.google.com/file/d/1MJyKp7X2xA_8PwSQOcKqc5MPCXZAyIkW/view?usp=sharing}. 
\end{enumerate}

\textbf{One-Class Classification (OCC) Methods.} These methods are trained exclusively on normal samples and detect anomalies as deviations from the learned normal distribution. 
\begin{enumerate}
    \item \textbf{OC-SVM}~\citep{scholkopf2001estimating} (One-Class Support Vector Machine) learns a decision boundary in a high-dimensional feature space that encloses the majority of normal training data. Test points falling outside this boundary are flagged as anomalies.
    \item\textbf{Autoencoder}~\citep{sakurada2014anomaly} projects data to a lower-dimensional latent space and reconstructs it through an encoding-decoding architecture. Anomalies are characterized by high reconstruction errors, as the model fails to accurately reconstruct patterns unseen during training.

    \item\textbf{DeepSVDD}~\citep{ruff2018deep} trains a neural network to map normal data into a compact hypersphere in a latent space. Anomaly scores are computed as the distance of a test point's representation from the center of the hypersphere.

    \item\textbf{TAM}~\citep{qiao2023truncated} learns node representations by maximizing local node affinity based on the one-class homophily property, where normal nodes exhibit stronger similarity to their neighbors. Anomalies are identified as nodes with low local affinity.

    \item\textbf{DOMINANT}~\citep{ding2019deep} leverages deep graph autoencoders to reconstruct both the graph structure and node attributes. Anomaly scores are derived from the joint reconstruction error of these two components.

    \item\textbf{BlindGuard}~\citep{BlindGuard} learns to detect anomalies via corruption-guided training, where pseudo anomalies are generated through feature perturbations and optimized using contrastive learning. 
\end{enumerate}

\noindent\textbf{Semi-Supervised Methods.} Require a small set of labeled abnormal data as the training set. 
\begin{enumerate}
    \item\textbf{DeepSAD}~\citep{ruff2019deep} extends the Deep SVDD framework to the semi-supervised setting by incorporating a small number of labeled anomalies during training. The objective encourages normal samples to map close to the hypersphere center while pushing labeled anomalies away from it.

    \item\textbf{DevNet}~\citep{pang2019deep} is a semi-supervised anomaly detection method that leverages a small set of labeled anomalies to learn a deviation network. It optimizes an anomaly score function such that labeled anomalies receive significantly higher scores than normal samples.

    \item\textbf{GGAD}~\citep{qiao2024generative} exploits a limited number of labeled anomalies within a graph-based framework to guide the anomaly detection process. It propagates anomaly information through the graph structure to improve detection coverage.

    \item\textbf{G-Safeguard (Semi-sup)}~\citep{gsafeguard} is the semi-supervised variant of G-Safeguard, where only a small subset of labeled anomalies is provided instead of the full set.

    \item\textbf{Gemma4-E2B (FS)}, \textbf{Gemma4-E4B (FS)}, \textbf{Gemma4-31B (FS)}, and \textbf{Gemma4-26B-A4B (FS)}~\citep{gemma4_deepmind_2026} are applied in a few-shot setting where 3 normal and 3 abnormal examples are provided in the prompt as context. This allows the model to better calibrate its notion of normality without any gradient-based training.  
\end{enumerate}

\noindent\textbf{Supervised Methods.} Require labeled normal and abnormal data as the training set. 
\begin{enumerate}
    \item \textbf{SVM}~\citep{hearst1998support} (Support Vector Machine) is a classical supervised classification method that finds a maximum-margin hyperplane to separate normal and anomalous samples in the feature space.

    \item\textbf{Random Forest}~\citep{breiman2001random} is an ensemble learning method that constructs multiple decision trees during training and outputs the majority class prediction. It is robust to overfitting and performs well on high-dimensional tabular data.

    \item\textbf{XGBoost}~\citep{chen2016xgboost} is a gradient boosting framework that builds an ensemble of decision trees in a sequential manner, where each tree corrects the errors of the previous one. It is widely adopted for its strong performance on structured data.

    \item\textbf{G-Safeguard}~\citep{gsafeguard} is a supervised graph-based anomaly detection method that models MAS as a graph and uses GNN-based anomaly detection on the multi-agent utterance topology to identify compromised agents.
\end{enumerate}

\textbf{Implementation Details.} Tabular methods are implemented via Scikit-learn~\citep{scikit-learn} and PyOD~\citep{PyOD_2019} with default hyperparameters; graph-based methods adapt author-provided codebases; and LLM-as-detector methods are motivated by MAST prompting~\citep{MAST}. All methods generate an anomaly score for each test sample. These scores are then converted into binary labels by thresholding based on the known number of anomalies in the test set, following prior work \citep{goad_2020, anomaly_ad_crtl_2022}. LLM's use logits of a constrained single-token (``yes''/``no'') output as anomaly scores.


\textbf{AD Evaluation Resource Cost.} The non-LLM detectors run on a single NVIDIA L40s GPU and complete within minutes per method (3--26 minute range), with tabular methods averaging 4.8 minutes, graph-based methods 9.0 minutes, and G-Safeguard 4.6 minutes, totaling under 2.4 hours. The LLM-based detectors run on a single NVIDIA A100 GPU. We benchmark four Gemma variants from 2B to 31B parameters; the largest (Gemma-4-31B) processes 69.94M tokens across 1{,}345 traces (33{,}000 tokens/trace under semi-supervised; 19{,}000 under unsupervised), taking 6.2 hours, while smaller variants (Gemma-4-E2B/E4B) take 2.5--3.0 hours each. Across all Gemma variants, total LLM-based benchmarking takes 14.0 hours.

\begin{figure}[h]
    \centering
        \vspace{-10pt}
    \includegraphics[width=0.5\linewidth]{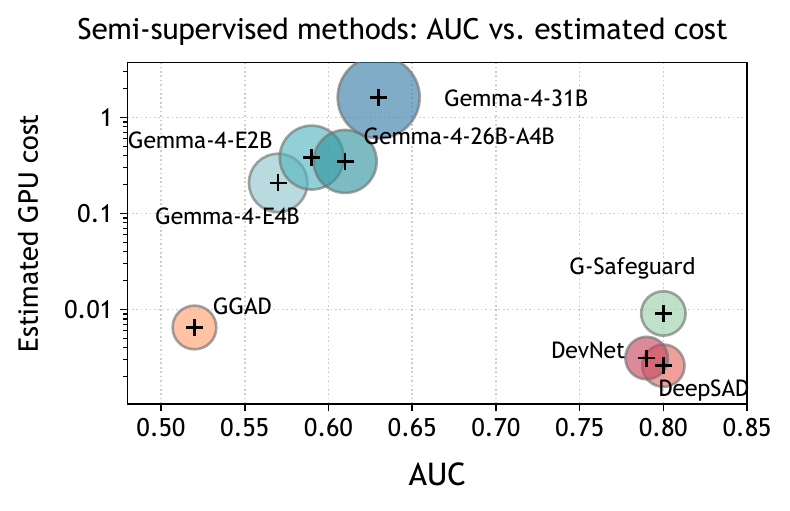}
    \vspace{-15pt}
    \caption{AUCROC vs. Normalized runtime GPU cost of semi-supervised methods.}
    \label{fig:cost_auc}
\end{figure}

\textbf{Detection Performance vs.\ Runtime Cost.}
Figure~\ref{fig:cost_auc} compares the detection performance and normalized runtime GPU cost of semi-supervised methods. LLM-based detectors incur orders-of-magnitude higher runtime cost while achieving only moderate AUC ($\sim0.57$--$0.63$). In contrast, several non-LLM methods, particularly DeepSAD, DevNet, and G-Safeguard, achieve substantially higher detection performance ($\sim0.79$--$0.80$) at much lower cost. This large efficiency gap suggests that scaling LLM-based detectors alone is not a cost-effective path for MAS AD, motivating lightweight, hybrid, or selectively invoked LLM-based designs.    
\section{Additional Evaluation Results}
\label{app:more_results}

\begin{table*}[ht]
    \centering
    \small
    \setlength{\tabcolsep}{3pt}
    \caption{
        Overall anomaly detection performance across four supervision settings. Dark blue cells indicate $p < 0.01$ and light blue cells indicate $p < 0.05$ relative to \textit{Random} performance.
    }
    \label{tab:main_results}
    \resizebox{\linewidth}{!}{
    \begin{tabular}{lllcccc}
\toprule
\textbf{Supervision} & \textbf{Method} & \textbf{Method Family} & \textbf{F1} & \textbf{Acc} & \textbf{AUC} & \textbf{Bal. Acc} \\
\midrule
\multirow{1}{*}{-} & Random & Random & $.176_{ \pm .021 }$ & $.705_{ \pm .021 }$ & $.494_{ \pm .013 }$ & $.498_{ \pm .013 }$  \\
\midrule
\multirow{7}{*}{Unsupervised} 
& Isolation Forest & Tabular & \cellcolor{blue!50} $.281_{ \pm .041 }$ & \cellcolor{blue!50} $.743_{ \pm .012 }$ & \cellcolor{blue!25} $.543_{ \pm .039 }$ & \cellcolor{blue!50} $.562_{ \pm .020 }$  \\
& KNN & Tabular & \cellcolor{blue!50} $.239_{ \pm .032 }$ & $.728_{ \pm .024 }$ & $.488_{ \pm .038 }$ & \cellcolor{blue!50} $.537_{ \pm .021 }$  \\
& LOF & Tabular & $.056_{ \pm .009 }$ & $.662_{ \pm .027 }$ & $.378_{ \pm .026 }$ & $.425_{ \pm .011 }$  \\
& Gemma4-E2B (zero-shot) & LLM & \cellcolor{blue!50} $.296_{ \pm .013 }$ & \cellcolor{blue!50} $.746_{ \pm .014 }$ & \cellcolor{blue!50} $.560_{ \pm .014 }$ & \cellcolor{blue!50} $.571_{ \pm .005 }$  \\
& Gemma4-E4B (zero-shot) & LLM & \cellcolor{blue!50} $.332_{ \pm .027 }$ & \cellcolor{blue!50} $.758_{ \pm .018 }$ & \cellcolor{blue!50} $.596_{ \pm .020 }$ & \cellcolor{blue!50} $.593_{ \pm .017 }$  \\
& Gemma4-26B-A4B (zero-shot) & LLM & \cellcolor{blue!50} $.244_{ \pm .035 }$ & \cellcolor{blue!25} $.729_{ \pm .013 }$ & \cellcolor{blue!50} $.542_{ \pm .021 }$ & \cellcolor{blue!50} $.540_{ \pm .017 }$  \\
& Gemma4-31B (zero-shot) & LLM & \cellcolor{blue!50} $.288_{ \pm .033 }$ & \cellcolor{blue!50} $.741_{ \pm .010 }$ & \cellcolor{blue!50} $.594_{ \pm .014 }$ & \cellcolor{blue!50} $.564_{ \pm .014 }$  \\
\midrule
\multirow{6}{*}{One-Class} 
& Autoencoder & Tabular & \cellcolor{blue!50} $.296_{ \pm .036 }$ & \cellcolor{blue!50} $.748_{ \pm .020 }$ & $.511_{ \pm .029 }$ & \cellcolor{blue!50} $.571_{ \pm .021 }$  \\
& DeepSVDD & Tabular & \cellcolor{blue!25} $.245_{ \pm .056 }$ & \cellcolor{blue!25} $.731_{ \pm .020 }$ & \cellcolor{blue!25} $.592_{ \pm .102 }$ & \cellcolor{blue!25} $.541_{ \pm .031 }$  \\
& OC-SVM & Tabular & \cellcolor{blue!50} $.293_{ \pm .034 }$ & \cellcolor{blue!50} $.747_{ \pm .018 }$ & \cellcolor{blue!50} $.600_{ \pm .024 }$ & \cellcolor{blue!50} $.570_{ \pm .019 }$  \\
& DOMINANT & Graph & \cellcolor{blue!50} $.269_{ \pm .030 }$ & \cellcolor{blue!25} $.739_{ \pm .011 }$ & \cellcolor{blue!50} $.702_{ \pm .010 }$ & \cellcolor{blue!50} $.555_{ \pm .012 }$  \\
& TAM & Graph & $.130_{ \pm .041 }$ & $.688_{ \pm .032 }$ & $.423_{ \pm .011 }$ & $.470_{ \pm .029 }$  \\
& BlindGuard & MAS-specific & $.126_{ \pm .029 }$ & $.687_{ \pm .027 }$ & $.415_{ \pm .015 }$ & $.467_{ \pm .021 }$  \\
\midrule
\multirow{8}{*}{Semi-Supervised} 
& DeepSAD & Tabular & \cellcolor{blue!50} $.490_{ \pm .057 }$ & \cellcolor{blue!50} $.818_{ \pm .009 }$ & \cellcolor{blue!50} $.803_{ \pm .016 }$ & \cellcolor{blue!50} $.690_{ \pm .030 }$  \\
& DevNet & Tabular & \cellcolor{blue!50} $.481_{ \pm .062 }$ & \cellcolor{blue!50} $.815_{ \pm .015 }$ & \cellcolor{blue!50} $.788_{ \pm .015 }$ & \cellcolor{blue!50} $.684_{ \pm .034 }$  \\
& GGAD & Graph & $.172_{ \pm .053 }$ & $.704_{ \pm .018 }$ & $.516_{ \pm .027 }$ & $.496_{ \pm .028 }$  \\
& G-Safe (semi-sup) & MAS-specific & \cellcolor{blue!50} $.324_{ \pm .080 }$ & \cellcolor{blue!50} $.841_{ \pm .013 }$ & \cellcolor{blue!50} $.797_{ \pm .018 }$ & \cellcolor{blue!50} $.600_{ \pm .038 }$  \\
& Gemma4-E2B (few-shot) & LLM & \cellcolor{blue!50} $.303_{ \pm .019 }$ & $.739_{ \pm .025 }$ & \cellcolor{blue!25} $.571_{ \pm .027 }$ & \cellcolor{blue!50} $.572_{ \pm .015 }$  \\
& Gemma4-E4B (few-shot) & LLM & \cellcolor{blue!50} $.342_{ \pm .039 }$ & $.753_{ \pm .033 }$ & \cellcolor{blue!25} $.591_{ \pm .043 }$ & \cellcolor{blue!25} $.596_{ \pm .030 }$  \\
& Gemma4-26B-A4B (few-shot) & LLM & \cellcolor{blue!50} $.312_{ \pm .041 }$ & $.745_{ \pm .031 }$ & \cellcolor{blue!50} $.613_{ \pm .020 }$ & \cellcolor{blue!25} $.578_{ \pm .029 }$  \\
& Gemma4-31B (few-shot) & LLM & \cellcolor{blue!50} $.348_{ \pm .012 }$ & \cellcolor{blue!25} $.759_{ \pm .026 }$ & \cellcolor{blue!25} $.628_{ \pm .038 }$ & \cellcolor{blue!50} $.600_{ \pm .016 }$  \\
\midrule
\multirow{4}{*}{Supervised} 
& SVM & Tabular & \cellcolor{blue!50} $.553_{ \pm .033 }$ & \cellcolor{blue!50} $.841_{ \pm .006 }$ & \cellcolor{blue!50} $.796_{ \pm .017 }$ & \cellcolor{blue!50} $.728_{ \pm .016 }$  \\
& Random Forest & Tabular & \cellcolor{blue!50} $\mathbf{.594_{ \pm .044 }}$ & \cellcolor{blue!50} $.855_{ \pm .007 }$ & \cellcolor{blue!50} $.863_{ \pm .007 }$ & \cellcolor{blue!50} $\mathbf{.754_{ \pm .023 }}$  \\
& XGBoost & Tabular & \cellcolor{blue!50} $.587_{ \pm .039 }$ & \cellcolor{blue!50} $.853_{ \pm .006 }$ & \cellcolor{blue!50} $\mathbf{.878_{ \pm .012 }}$ & \cellcolor{blue!50} $.749_{ \pm .020 }$  \\
& G-Safeguard & MAS-specific & \cellcolor{blue!50} $.551_{ \pm .030 }$ & \cellcolor{blue!50} $\mathbf{.855_{ \pm .012 }}$ & \cellcolor{blue!50} $.834_{ \pm .013 }$ & \cellcolor{blue!50} $.716_{ \pm .012 }$  \\
\bottomrule
    \end{tabular}}
\end{table*}

\begin{table*}[ht]
    \centering
    \small
    \setlength{\tabcolsep}{3pt}
    \caption{
        One-tailed two-sample t-test p-values (Method $>$ Random) for overall anomaly detection performance across four performance metrics.
    }
    \label{tab:main_results_pvalues_n5}
    \begin{tabular}{lllcccc}
\toprule
\textbf{Supervision} & \textbf{Method} & \textbf{Method Family} & \textbf{F1} & \textbf{Acc} & \textbf{AUC} & \textbf{Bal. Acc} \\
\midrule
\multirow{1}{*}{-} & Random & Random & - & - & - & -  \\
\midrule
\multirow{7}{*}{Unsupervised} 
& Isolation Forest & Tabular & $.001$ & $.006$ & $.022$ & $<.001$  \\
& KNN & Tabular & $.004$ & $.081$ & $.631$ & $.006$  \\
& LOF & Tabular & $1.000$ & $.988$ & $1.000$ & $1.000$  \\
& Gemma4-E2B (zero-shot) & LLM & $<.001$ & $.005$ & $<.001$ & $<.001$  \\
& Gemma4-E4B (zero-shot) & LLM & $<.001$ & $.002$ & $<.001$ & $<.001$  \\
& Gemma4-26B-A4B (zero-shot) & LLM & $.004$ & $.039$ & $.002$ & $.002$  \\
& Gemma4-31B (zero-shot) & LLM & $<.001$ & $.009$ & $<.001$ & $<.001$  \\
\midrule
\multirow{6}{*}{One-Class} 
& Autoencoder & Tabular & $<.001$ & $.006$ & $.144$ & $<.001$  \\
& DeepSVDD & Tabular & $.024$ & $.045$ & $.049$ & $.017$  \\
& OC-SVM & Tabular & $<.001$ & $.005$ & $<.001$ & $<.001$  \\
& DOMINANT & Graph & $<.001$ & $.010$ & $<.001$ & $<.001$  \\
& TAM & Graph & $.968$ & $.822$ & $1.000$ & $.952$  \\
& BlindGuard & MAS-specific & $.992$ & $.861$ & $1.000$ & $.986$  \\
\midrule
\multirow{8}{*}{Semi-Supervised} 
& DeepSAD & Tabular & $<.001$ & $<.001$ & $<.001$ & $<.001$  \\
& DevNet & Tabular & $<.001$ & $<.001$ & $<.001$ & $<.001$  \\
& GGAD & Graph & $.562$ & $.528$ & $.079$ & $.564$  \\
& G-Safe (semi-sup) & MAS-specific & $.006$ & $<.001$ & $<.001$ & $.001$  \\
& Gemma4-E2B (few-shot) & LLM & $<.001$ & $.064$ & $.014$ & $.001$  \\
& Gemma4-E4B (few-shot) & LLM & $.005$ & $.055$ & $.027$ & $.011$  \\
& Gemma4-26B-A4B (few-shot) & LLM & $.009$ & $.066$ & $.001$ & $.015$  \\
& Gemma4-31B (few-shot) & LLM & $<.001$ & $.021$ & $.010$ & $<.001$  \\
\midrule
\multirow{4}{*}{Supervised} 
& SVM & Tabular & $<.001$ & $<.001$ & $<.001$ & $<.001$  \\
& Random Forest & Tabular & $\mathbf{<.001}$ & $<.001$ & $<.001$ & $\mathbf{<.001}$  \\
& XGBoost & Tabular & $<.001$ & $<.001$ & $\mathbf{<.001}$ & $<.001$  \\
& G-Safeguard & MAS-specific & $<.001$ & $\mathbf{<.001}$ & $<.001$ & $<.001$  \\
\bottomrule
    \end{tabular}
\end{table*}

\begin{table*}[ht]
    \centering
    \scriptsize
    \setlength{\tabcolsep}{3pt}
    \caption{
        AUCROC per (method $\times$ action), mean$\,\pm\,$std over available train-test splits. Action success rate shown in parentheses. Per-column maxima in \textbf{bold}
    }
    \label{tab:aucroc-by-action}
    \begin{tabular}{llccccccc|c}
        \toprule
        Family & Method & SP & OC & SC & OI & TC & AD & OP & Avg \\
         & & (87.0\%) & (96.5\%) & (76.0\%) & (91.9\%) & (62.5\%) & (57.3\%) & (99.5\%) & \\
        \midrule
        \multirow{7}{*}{Unsupervised} & iForest & 0.49\,$\pm$\,0.13 & 0.54\,$\pm$\,0.02 & 0.60\,$\pm$\,0.06 & 0.56\,$\pm$\,0.04 & 0.51\,$\pm$\,0.04 & 0.53\,$\pm$\,0.07 & 0.47\,$\pm$\,0.24 & 0.53\,$\pm$\,0.05 \\
         & kNN & 0.37\,$\pm$\,0.17 & 0.52\,$\pm$\,0.07 & 0.68\,$\pm$\,0.09 & 0.58\,$\pm$\,0.04 & 0.54\,$\pm$\,0.09 & 0.61\,$\pm$\,0.03 & 0.52\,$\pm$\,0.18 & 0.55\,$\pm$\,0.10 \\
         & LOF & 0.71\,$\pm$\,0.16 & 0.49\,$\pm$\,0.05 & 0.62\,$\pm$\,0.07 & 0.52\,$\pm$\,0.05 & 0.57\,$\pm$\,0.08 & 0.59\,$\pm$\,0.03 & 0.56\,$\pm$\,0.23 & 0.58\,$\pm$\,0.07 \\
         & Gemma4-E2B & 0.24\,$\pm$\,0.10 & 0.29\,$\pm$\,0.08 & 0.62\,$\pm$\,0.02 & 0.71\,$\pm$\,0.05 & 0.40\,$\pm$\,0.07 & 0.61\,$\pm$\,0.02 & 0.41\,$\pm$\,0.27 & 0.47\,$\pm$\,0.18 \\
         & Gemma4-E4B & 0.19\,$\pm$\,0.03 & 0.46\,$\pm$\,0.06 & 0.63\,$\pm$\,0.02 & 0.65\,$\pm$\,0.03 & 0.53\,$\pm$\,0.05 & 0.54\,$\pm$\,0.04 & 0.72\,$\pm$\,0.11 & 0.53\,$\pm$\,0.17 \\
         & Gemma4-26B & 0.23\,$\pm$\,0.04 & 0.51\,$\pm$\,0.09 & 0.64\,$\pm$\,0.02 & 0.56\,$\pm$\,0.03 & 0.67\,$\pm$\,0.03 & 0.59\,$\pm$\,0.04 & 0.55\,$\pm$\,0.24 & 0.54\,$\pm$\,0.14 \\
         & Gemma4-31B & 0.36\,$\pm$\,0.06 & 0.65\,$\pm$\,0.09 & \textbf{0.77\,$\pm$\,0.02} & 0.65\,$\pm$\,0.02 & 0.68\,$\pm$\,0.04 & 0.65\,$\pm$\,0.02 & 0.74\,$\pm$\,0.05 & 0.64\,$\pm$\,0.13 \\
        \midrule
        \multirow{6}{*}{OCC} & AutoEncoder & 0.59\,$\pm$\,0.15 & 0.54\,$\pm$\,0.05 & 0.68\,$\pm$\,0.08 & 0.58\,$\pm$\,0.04 & 0.62\,$\pm$\,0.09 & 0.60\,$\pm$\,0.06 & 0.53\,$\pm$\,0.21 & 0.59\,$\pm$\,0.05 \\
         & DeepSVDD & 0.48\,$\pm$\,0.18 & 0.54\,$\pm$\,0.05 & 0.62\,$\pm$\,0.04 & 0.51\,$\pm$\,0.05 & 0.50\,$\pm$\,0.11 & 0.49\,$\pm$\,0.03 & 0.59\,$\pm$\,0.19 & 0.53\,$\pm$\,0.05 \\
         & OC-SVM & 0.72\,$\pm$\,0.10 & 0.57\,$\pm$\,0.04 & 0.66\,$\pm$\,0.09 & 0.58\,$\pm$\,0.04 & 0.59\,$\pm$\,0.10 & 0.55\,$\pm$\,0.04 & 0.58\,$\pm$\,0.18 & 0.61\,$\pm$\,0.06 \\
         & DOMINANT & 0.26\,$\pm$\,0.12 & 0.51\,$\pm$\,0.04 & 0.59\,$\pm$\,0.06 & 0.57\,$\pm$\,0.05 & 0.37\,$\pm$\,0.03 & 0.33\,$\pm$\,0.05 & 0.50\,$\pm$\,0.23 & 0.45\,$\pm$\,0.13 \\
         & TAM & 0.14\,$\pm$\,0.06 & 0.29\,$\pm$\,0.08 & 0.46\,$\pm$\,0.09 & 0.50\,$\pm$\,0.07 & 0.46\,$\pm$\,0.08 & 0.54\,$\pm$\,0.04 & 0.72\,$\pm$\,0.15 & 0.44\,$\pm$\,0.19 \\
         & BlindGuard & 0.36\,$\pm$\,0.06 & 0.48\,$\pm$\,0.09 & 0.37\,$\pm$\,0.09 & 0.51\,$\pm$\,0.04 & 0.46\,$\pm$\,0.07 & 0.49\,$\pm$\,0.04 & 0.62\,$\pm$\,0.26 & 0.47\,$\pm$\,0.09 \\
        \midrule
        \multirow{8}{*}{Semi-supervised} & DeepSAD & 0.85\,$\pm$\,0.11 & 0.67\,$\pm$\,0.06 & 0.68\,$\pm$\,0.07 & 0.64\,$\pm$\,0.03 & 0.64\,$\pm$\,0.04 & 0.76\,$\pm$\,0.03 & 0.56\,$\pm$\,0.31 & 0.69\,$\pm$\,0.09 \\
         & DevNet & 0.80\,$\pm$\,0.10 & 0.57\,$\pm$\,0.05 & 0.64\,$\pm$\,0.08 & 0.60\,$\pm$\,0.07 & 0.62\,$\pm$\,0.03 & 0.69\,$\pm$\,0.07 & 0.60\,$\pm$\,0.13 & 0.65\,$\pm$\,0.08 \\
         & GGAD & 0.51\,$\pm$\,0.16 & 0.49\,$\pm$\,0.08 & 0.53\,$\pm$\,0.02 & 0.49\,$\pm$\,0.05 & 0.52\,$\pm$\,0.09 & 0.54\,$\pm$\,0.08 & 0.48\,$\pm$\,0.34 & 0.51\,$\pm$\,0.02 \\
         & G-Safeguard & 0.62\,$\pm$\,0.09 & 0.63\,$\pm$\,0.08 & 0.63\,$\pm$\,0.03 & 0.70\,$\pm$\,0.03 & 0.73\,$\pm$\,0.02 & 0.74\,$\pm$\,0.02 & 0.65\,$\pm$\,0.24 & 0.67\,$\pm$\,0.05 \\
         & Gemma4-E2B & 0.44\,$\pm$\,0.04 & 0.34\,$\pm$\,0.08 & 0.62\,$\pm$\,0.05 & 0.64\,$\pm$\,0.05 & 0.40\,$\pm$\,0.11 & 0.60\,$\pm$\,0.03 & 0.51\,$\pm$\,0.28 & 0.51\,$\pm$\,0.12 \\
         & Gemma4-E4B & 0.32\,$\pm$\,0.05 & 0.54\,$\pm$\,0.09 & 0.66\,$\pm$\,0.05 & 0.61\,$\pm$\,0.04 & 0.57\,$\pm$\,0.10 & 0.58\,$\pm$\,0.07 & 0.76\,$\pm$\,0.26 & 0.58\,$\pm$\,0.13 \\
         & Gemma4-26B & 0.36\,$\pm$\,0.05 & 0.67\,$\pm$\,0.02 & 0.65\,$\pm$\,0.03 & 0.66\,$\pm$\,0.06 & 0.69\,$\pm$\,0.04 & 0.60\,$\pm$\,0.01 & 0.77\,$\pm$\,0.18 & 0.63\,$\pm$\,0.13 \\
         & Gemma4-31B & 0.42\,$\pm$\,0.08 & 0.85\,$\pm$\,0.03 & 0.74\,$\pm$\,0.05 & 0.78\,$\pm$\,0.01 & 0.65\,$\pm$\,0.10 & 0.65\,$\pm$\,0.03 & \textbf{0.89\,$\pm$\,0.08} & 0.71\,$\pm$\,0.16 \\
        \midrule
        \multirow{4}{*}{Supervised} & SVM & 0.82\,$\pm$\,0.11 & 0.63\,$\pm$\,0.06 & 0.73\,$\pm$\,0.08 & 0.56\,$\pm$\,0.04 & 0.73\,$\pm$\,0.05 & 0.76\,$\pm$\,0.03 & 0.68\,$\pm$\,0.14 & 0.70\,$\pm$\,0.09 \\
         & Rand.Forest & 0.84\,$\pm$\,0.11 & 0.78\,$\pm$\,0.07 & 0.74\,$\pm$\,0.06 & 0.73\,$\pm$\,0.03 & 0.70\,$\pm$\,0.07 & 0.79\,$\pm$\,0.02 & 0.58\,$\pm$\,0.30 & 0.74\,$\pm$\,0.08 \\
         & XGBoost & \textbf{0.87\,$\pm$\,0.08} & \textbf{0.89\,$\pm$\,0.04} & 0.72\,$\pm$\,0.06 & \textbf{0.81\,$\pm$\,0.03} & 0.72\,$\pm$\,0.05 & 0.79\,$\pm$\,0.02 & 0.65\,$\pm$\,0.33 & \textbf{0.78\,$\pm$\,0.09} \\
         & G-Safeguard & 0.71\,$\pm$\,0.06 & 0.65\,$\pm$\,0.07 & 0.61\,$\pm$\,0.06 & 0.73\,$\pm$\,0.02 & \textbf{0.78\,$\pm$\,0.03} & \textbf{0.79\,$\pm$\,0.03} & 0.71\,$\pm$\,0.15 & 0.71\,$\pm$\,0.07 \\
        \midrule
        \multicolumn{2}{l}{\textit{Avg}} & 0.51\,$\pm$\,0.23 & 0.56\,$\pm$\,0.15 & 0.64\,$\pm$\,0.09 & 0.62\,$\pm$\,0.09 & 0.59\,$\pm$\,0.11 & 0.62\,$\pm$\,0.11 & 0.61\,$\pm$\,0.11 & 0.59\,$\pm$\,0.04 \\
        \bottomrule
    \end{tabular}
\end{table*}

\begin{figure*}[ht]
    \centering
    \includegraphics[width=\linewidth]{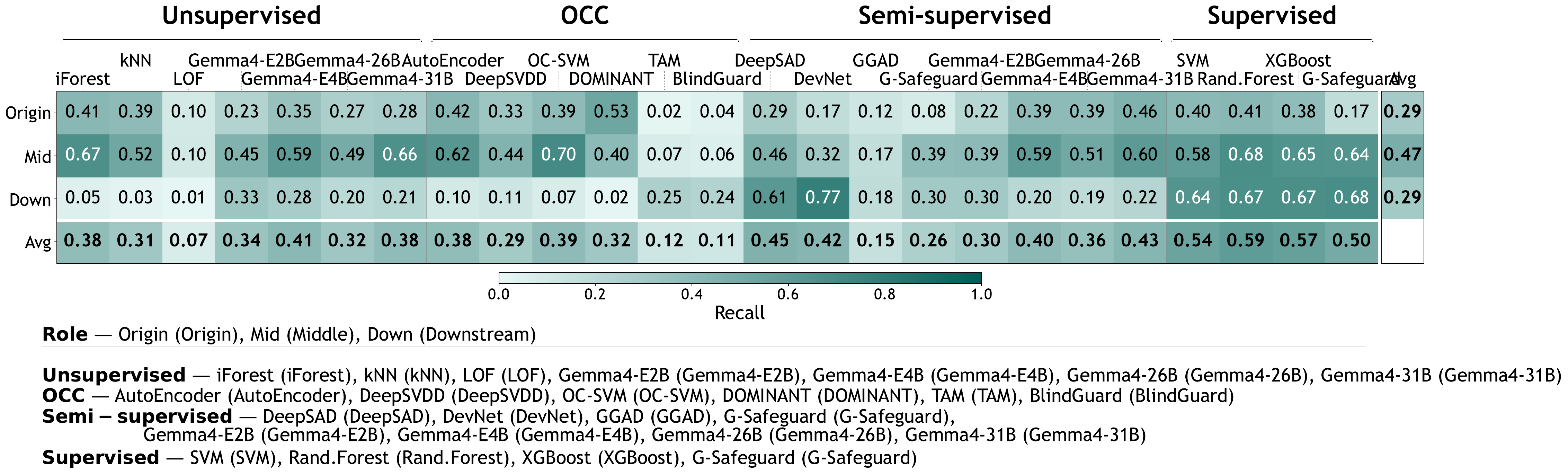}
    \caption{
        Per-method recall on multi-anomaly propagation chains, broken down by step role. \textbf{Rows}: step role, including \textbf{Origin} (the first anomalous step), \textbf{Mid} (intermediate anomalous steps), and \textbf{Down} (the last anomalous step); \textbf{Bottom row}: per-method average across roles, indicating overall propagation detection performance. \textbf{Columns}: AD methods grouped by supervision level. \textbf{Rightmost column}: per-role average across methods. Middle steps have the highest average recall across methods ($0.47$), while origin and downstream steps are substantially harder ($0.29$ and $0.29$, respectively). Overall, Random Forest achieves the best average propagation-role recall ($0.59$), followed by XGBoost ($0.57$) and SVM ($0.54$). Per role, DOMINANT performs best on origin steps (recall $0.53$), OC-SVM performs best on middle steps (recall $0.70$), and DevNet performs best on downstream steps (recall $0.77$).}
        \label{fig:propagate_deteailed}
\end{figure*}



\end{document}